\documentclass[letterpaper, 10 pt, conference]{ieeeconf}  

\IEEEoverridecommandlockouts     
\usepackage[T1]{fontenc}
\usepackage{bbm}
\usepackage{etoolbox}
\usepackage{multicol}
\usepackage{cite}
\usepackage[bookmarks=true]{hyperref}
\usepackage{graphics} 
\usepackage{epsfig}
\usepackage{subcaption}
\usepackage{amsmath}
\usepackage{amssymb} 
\usepackage{mathrsfs}
\usepackage{amsfonts}
\usepackage{svg}
\usepackage{wrapfig}
\usepackage{subcaption}
\usepackage{graphicx}
\usepackage{float}
\usepackage{ctable}
\usepackage{cuted}
\usepackage{colortbl}
\usepackage{multirow}
\usepackage{algorithm}
\usepackage{algpseudocode}
\usepackage{xcolor}
\usepackage[normalem]{ulem}
\usepackage{soul}
\usepackage{contour}   
\usepackage{pifont}
\usepackage{setspace}
\usepackage{threeparttable}
\usepackage{bm} 
\usepackage{xspace}
\usepackage{wasysym}

\definecolor{formalgreen}{rgb}{0.1, 0.7, 0.1}
\definecolor{formalred}{rgb}{0.9, 0.2, 0.2}

\newif\ifRRMode
\RRModefalse

\newcommand{\deleted}[1]{\ifRRMode\textcolor{red}{\sout{#1}}\else\relax\fi}
\newcommand{\added}[1]{\ifRRMode\textcolor{blue}{\uwave{#1}}\else#1\fi}

\setul{1pt}{1pt}
\setuldepth{content}

\newcommand{\TODO}[1][]{\textcolor{red}{\bf [TODO]}}

\title{\LARGE \bf
TelePreview: A User-Friendly Teleoperation System with \\
Virtual Arm Assistance for Enhanced Effectiveness
}

\author{Jingxiang Guo$^{1*}$, Jiayu Luo$^{1*}$, Zhenyu Wei$^{1*}$, Yiwen Hou$^{1}$,\\
Zhixuan Xu$^{1}$, Xiaoyi Lin$^{1}$, Chongkai Gao$^{1}$, Lin Shao$^{1}$\textsuperscript{\textdagger}%
\thanks{* denotes equal contribution}%
\thanks{\textdagger \ denotes the corresponding author}%
\thanks{$^{1}$\texttt{Jingxiang Guo, Jiayu Luo, Zhenyu Wei, Yiwen Hou, Zhixuan Xu, Xiaoyi Lin, Chongkai Gao, and Lin Shao are with the Department of Computer Science, National University of Singapore. 
        {\tt\small jingxiangguo@u.nus.edu, jiayu@comp.nus.edu.sg, Zhenyu\_Wei@sjtu.edu.cn, yiwen328@comp.nus.edu.sg, zhixuanxu@u.nus.edu, gaochongkai@u.nus.edu, linshao@nus.edu.sg}}
}
}

\begin{document}
\maketitle
\thispagestyle{empty}
\pagestyle{empty}

\begin{strip}
    \vspace{-3.5cm}
    \begin{figure}[H]
        \centering
\includegraphics[width=\textwidth]{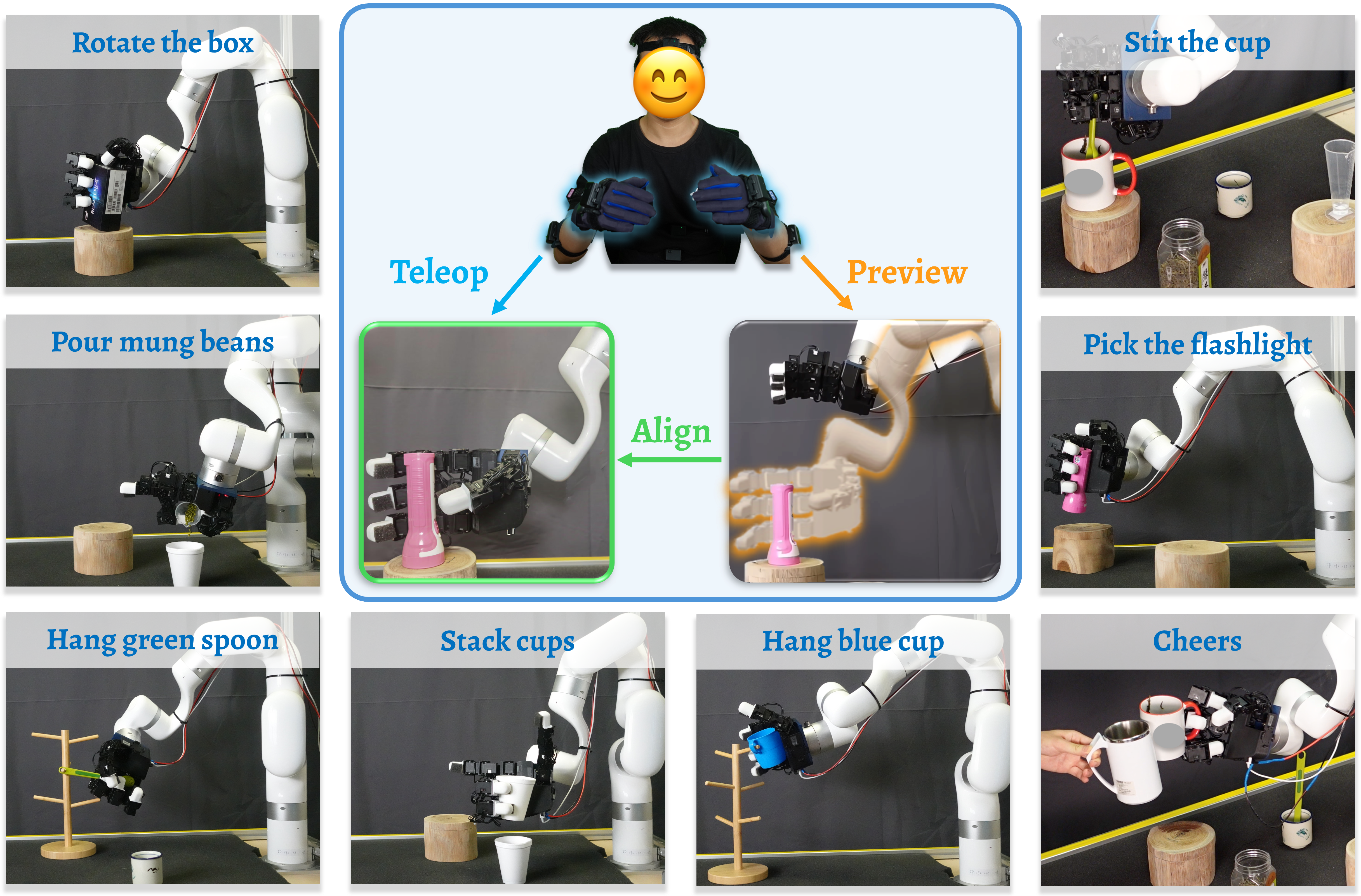}
        \caption{\parbox{0.95\textwidth}        {\textbf{TelePreview} is a user-friendly teleoperation system enabling the real-time virtual preview before robot execution.}}
        \label{fig:teaser}
        \vspace{-1cm}
    \end{figure}
\end{strip}

\begin{abstract}
Teleoperation provides an effective way to collect robot data, which is crucial for learning from demonstrations. In this field, teleoperation faces several key challenges: user-friendliness for new users, safety assurance, and transferability across different platforms. While collecting real robot dexterous manipulation data by teleoperation to train robots has shown impressive results on diverse tasks, due to the morphological differences between human and robot hands, it is not only hard for new users to understand the action mapping but also raises potential safety concerns during operation.
To address these limitations, we introduce TelePreview. This teleoperation system offers real-time visual feedback on robot actions based on human user inputs, with a total hardware cost of less than \$1,000. TelePreview allows the user to see a virtual robot that represents the outcome of the user's next movement. By enabling flexible switching between command visualization and actual execution, this system helps new users learn how to demonstrate quickly and safely.
We demonstrate that it outperforms other teleoperation systems across five tasks, emphasize its ease of use, and highlight its straightforward deployment across diverse robotic platforms.
We release our code and a deployment document on our website 
\url{https://nus-lins-lab.github.io/telepreview-web/}
\end{abstract}

\section{Introduction}
\begin{figure*}[htbp]
  \centering
  \vspace{10px}
  \includegraphics[width=\textwidth]{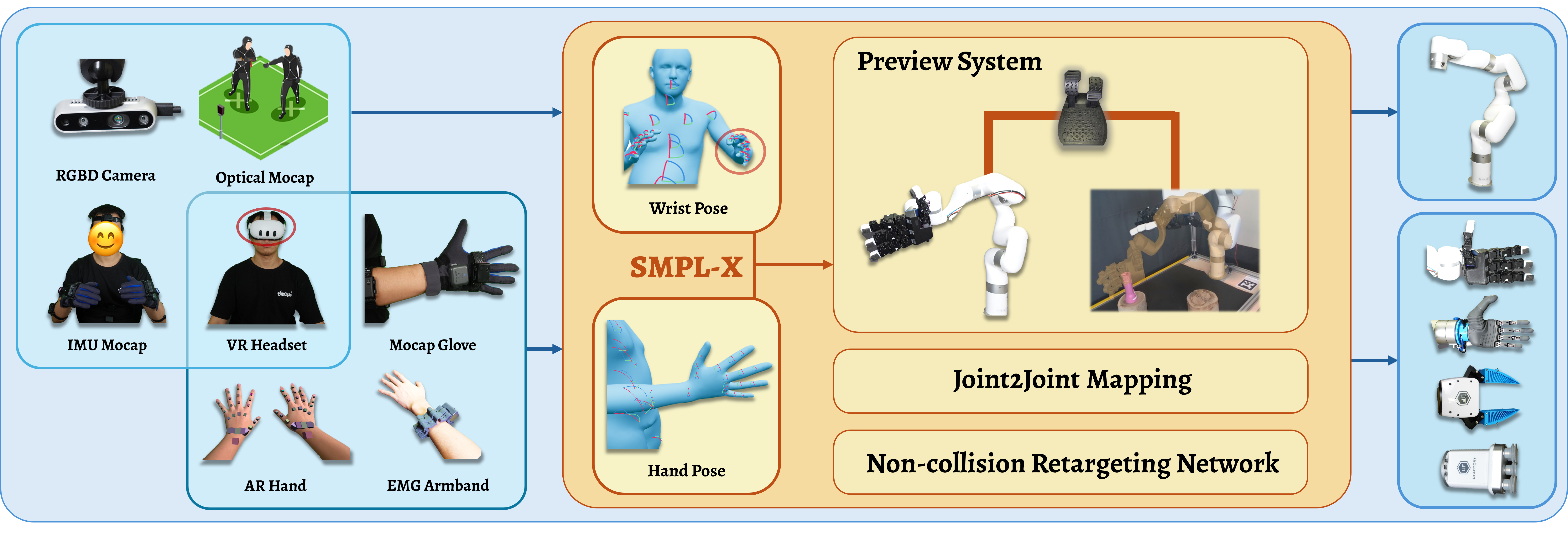}
  \caption{\textbf{Overview of the \deleted{TelePreview }System Architecture} \deleted{The system includes three components: }(1) Various input devices for capturing human motion\deleted{including RGB-D cameras, IMU mocap suits, VR headsets, mocap gloves, AR hand tracking, and EMG armbands}\added{. Only one device from each of the two input groups (wrist pose and hand gesture) is required during operation. The VR headset is shown at the intersection of the two groups, as it can capture both}; (2) \deleted{A central processing pipeline that utilizes SMPL-X as a unified representation to process wrist and hand pose data, feeding into our Preview System via joint-to-joint mapping and a non-collision retargeting network} \added{A processing pipeline based on SMPL-X that performs joint mapping and collision-free retargeting}; (3) \deleted{Support for different robot platforms as output devices for executing the mapped movements} \added{Output to various robot platforms for executing the mapped motions}.
}
  \label{fig: framework}
  \vspace{-0.5cm}
\end{figure*}

Researchers have long recognized teleoperation as an essential component for gathering on-robot data used in learning from demonstrations~\cite{iyer2024open, fu2024mobile, qin2023anyteleop, shaw2024bimanual, park2024using, wang2024dexcap}. An intuitive and user-friendly teleoperation system is crucial for collecting high-quality, diverse, and scalable data. However, effectively controlling robots with dexterous hands remains a significant challenge. The fundamental differences between human and robotic hands make the direct mapping of human movements to robotic actions exceptionally complex~\cite{cheng2024open, kennel2021manipulability, huang2025human}. These inherent structural and functional disparities, combined with the precision required for fine-grained tasks, often result in inefficient and potentially compromised data collection processes.

Existing teleoperation systems face three significant challenges. First, new users often struggle to control the robot effectively due to the lack of intuitive feedback on how their commands translate to robot actions, especially during complex manipulation tasks. Second, without proper safeguards and real-time guidance, users may inadvertently issue unsafe commands that could damage the robot or its surroundings. Third, most current systems are tightly coupled with specific input devices or robot platforms, limiting their adaptability. These challenges significantly impact the quality and efficiency of data collection for robot learning.

To address these issues, we introduce \textbf{TelePreview}, a teleoperation system that delivers precise control while maintaining affordability through its sub-\$1,000 hardware setup, as shown in Fig. \ref{fig: framework}. At its core, TelePreview provides an interface that overlays a \textbf{virtual robotic arm} onto the real-world scene, enabling users to preview motions. Precisely, a foot pedal or other external signal IO allows for mode switching between \emph{preview} (virtual-only) and \emph{align} (physical action alignment) modes. In \textit{preview} mode, users can verify and refine their intended actions with a virtual robot arm that is spatially aligned with the physical robot, ensuring that the final commands are safe and accurate once applied in the real world. 

We present the significant contributions of TelePreview to advance teleoperation:

\begin{itemize}
    \item \textbf{Interactive Visual Assistance:} We develop a visualization interface that enables users to preview intended actions before execution, enhancing teleoperation precision and accessibility for both new and experienced users during complex tasks.
    \item \textbf{Low-Cost System:} We introduce an economical yet high-performance teleoperation solution that integrates inertial motion capture technology and mocap gloves at a total hardware cost of under \$1,000, making it affordable for research.
    \item \textbf{Easy Adaptation to New Hardware:} We design our approach so that migrating to different input or output devices requires adjusting only a small set of physically interpretable parameters, ensuring robust adaptation across diverse hardware setups.
\end{itemize}

\section{Related Work}
\begin{figure*}[tbp]
    \centering
    \vspace{10px}
    \includegraphics[width=\textwidth]{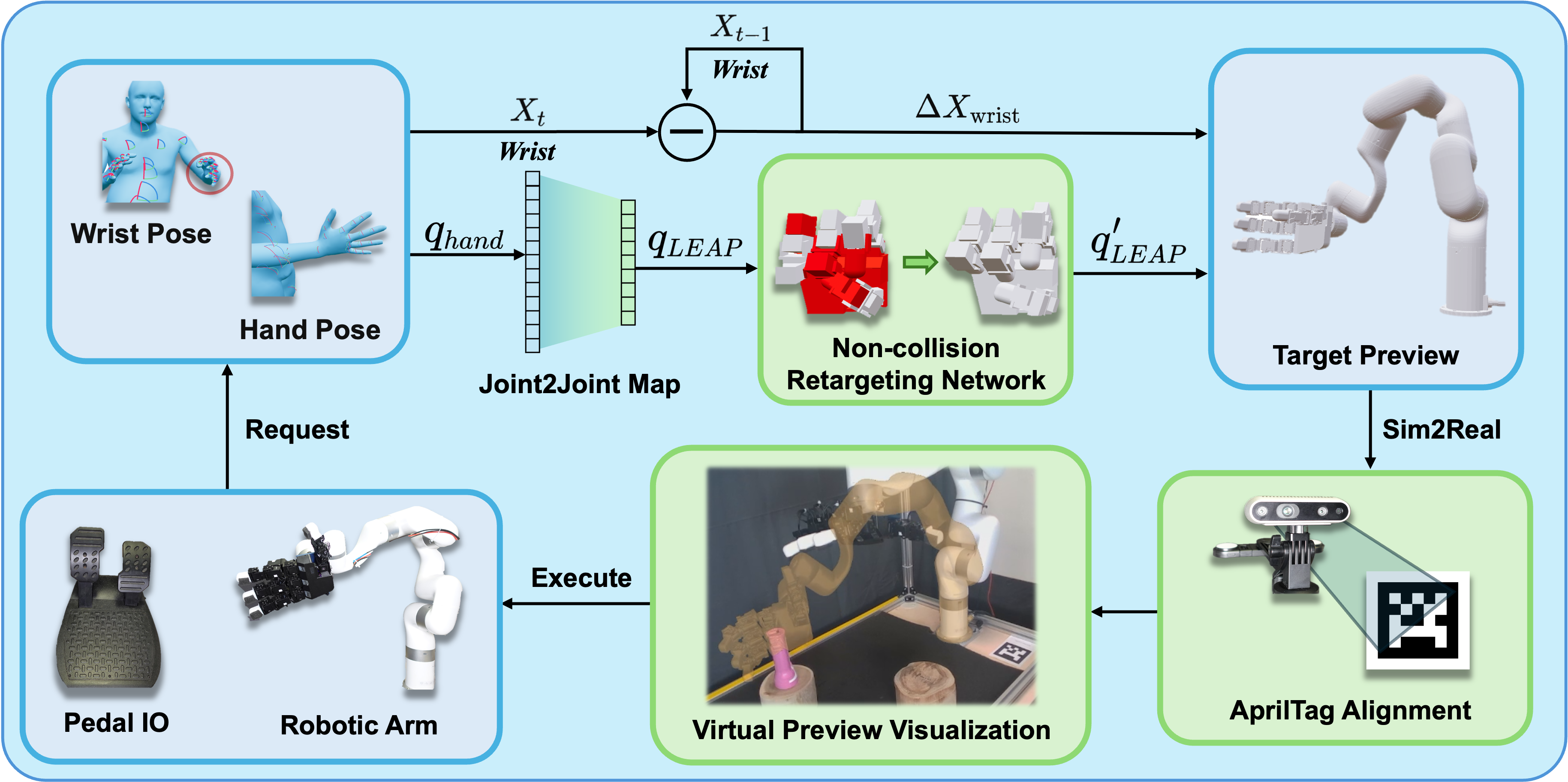}
    \caption{\textbf{Pipeline of Our Modules:} The system tracks user wrist and hand poses, maps them to robot joint configurations through joint-to-joint mapping and non-collision retargeting, and provides a visual preview before physical execution. We achieve precise alignment between virtual and physical robots through AprilTag calibration.}
    \label{fig: pipeline}
    \vspace{-0.5cm}
\end{figure*}

\subsection{\deleted{Robot} Teleoperation Frameworks for \added{Dexterous} Manipulation}

\deleted{The growing interest in training robots through imitation learning has driven the need for extensive, high-quality real robot datasets.} Teleoperation provides an efficient method for demonstrating and recording intricate robotic tasks\cite{darvish2023teleoperation, lichiardopol2007survey, hokayem2006bilateral}. \added{But Many existing teleoperation frameworks focus on low-degree-of-freedom (low-DoF) robotic grippers, using input modalities such as keyboards}\cite{katyal2014approaches}, \added{joysticks}\cite{aronson2018eye, scherzinger2023learning}\added{, or gamepads}\cite{micire2011design}\added{ to issue discrete commands. While these are effective for simple tasks, they are not sufficient for the fine-grained control of dexterous robotic hands with High DoFs.} 

Researchers have explored various teleoperation systems \added{for dexterous manipulation}, including VR headsets \cite{cheng2024open, ding2024bunny, iyer2024open}, motion capture \cite{9035064, stanton2012teleoperation, miller2004motion}, wearable gloves \cite{liu2019high, wang2024dexcap, mosbach2022accelerating, schwarz2021nimbro, audonnet2024telesim}, exoskeletons \cite{yang2024ace, wu2023gello}\deleted{, each offering distinct benefits in terms of accessibility, precision, and generalizability}.

\added{Using VR headsets}, vision-based teleoperation systems \cite{qin2023anyteleop, ding2024bunny, iyer2024open, cheng2024open} typically employ RGB\deleted{ or RGB-D} cameras to detect hand poses. \deleted{The hand pose retargeting module maps the human hand pose data obtained from perception algorithms into joint positions of the dexterous hand. We formulate this process as an optimization problem that minimizes the difference between the key point vectors of the human hand and the dexterous hand.}\added{These systems often rely on estimated key points from visual perception to map user motions onto the robot, typically using optimization-based retargeting methods.} 
\deleted{To generate the end-effector motions, 3D positions of key points in the camera frame are calculated by using an RGB-D camera. The wrist pose is then computed by keypoint positions in the local wrist frame and global camera frame using the Perspective-n-Point (PnP) algorithm.}

Although vision-based teleoperation systems offer a relatively low-cost solution, they face significant challenges, including limited degrees of freedom (DoFs), constrained reachable workspaces, and high computational demands.
\deleted{This stems from the requirement for operators to keep their hands within the camera's field of view, thereby restricting the effective action space. Although increasing the scale factor can amplify end-effector movement and expand the action space of the robot, it aggravates the difficulty for operators in performing fine-grained tasks.} These systems also struggle with occlusions, finger overlap, and other perception issues. The keypoint-based retargeting step can be computationally expensive in real time. Although specialized hardware \cite{fu2024mobile} can improve accuracy, it substantially increases system costs. 

In contrast, exoskeleton systems\cite{yang2024ace, wu2023gello} provide highly accurate motion capture with low latency but tend to be tightly coupled to specific hardware platforms, which limits their generalizability across different robot embodiments.

\added{In this work, we adopt a glove-based teleoperation framework that captures human hand motion at the joint level and maps it to a unified body model (e.g., SMPL-X) before retargeting it to the robot. This abstraction layer enables real-time, joint-level control of dexterous robotic hands while improving generalizability across different robot embodiments.} \deleted{Our proposed system directly uses the joint pose obtained from the motion capture suit and data gloves, preventing the deviation of vision-based systems and reducing the computational cost. This allows for real-time, joint-level control of dexterous hands without relying on visual tracking or external calibration.}\added{Although glove-based teleoperation has previously been considered expensive, recent commercialization and the availability of low-cost, off-the-shelf data gloves have made this approach increasingly accessible.}

\deleted{Recent unified teleoperation frameworks attempt to address these limitations but often introduce new trade-offs. Some approaches prioritize implementation simplicity by tracking only fingertip positions rather than enabling detailed joint-level control. Furthermore, their customization options typically require time-consuming manual parameter tuning that requires significant expertise. }

\subsection{Visual Feedback in Teleoperation Systems}

Visual feedback plays a crucial role in teleoperation systems by providing users with spatial awareness and task-relevant information. Traditional approaches primarily rely on real-time camera feeds \cite{lawrence1993stability, ousaid2015stable}, which can be limited by occlusions and restricted viewing angles. Moreover, the lack of preview capabilities for intended commands leaves users (especially new users) uncertain about the robot's response, which can increase error rates and user fatigue.

To address these limitations, researchers have explored enhanced visualization techniques \cite{zhao2017augmented, lee2018implementation, milgram1995telerobotic, sivakumar2022robotic} \added{in AR-based teleoperation systems}\cite{wang2024eve, duan2023ar2}, \added{which aim to improve user understanding of spatial constraints and planned motions.} \deleted{Even though these approaches improve user perception by displaying planned trajectories and environmental constraints, they primarily design and visualize AR objects for a specific task, such as a pink box with dimensions of 10 cm × 10 cm × 10 cm. Any changes to the object's attributes (e.g., dimensions, color, or type) invalidate the preconfigured AR object, so the system must rebuild it to match the new characteristics.}

\deleted{Instead of visualizing the manipulated objects, our system provides an AR representation of the robot that directly maps operator commands. This design ensures safety for new users who may be unfamiliar with how the robot responds to their inputs. Our configuration allows users to use their own URDF file, eliminating the need to reconstruct manipulated objects for every new task.}

\added{Prior AR-based teleoperation systems typically display a virtual robot arm in the environment but do not spatially align it with the physical robot. As a result, they cannot capture interaction data during contact with real objects or the robot's behavior following physical execution. In contrast, our system overlays the AR robot directly on top of the physical robot, enabling users to first control a virtual preview and then switch to executing the same motion with the real hardware. This design allows us to capture the full sequence of a teleoperation task—from intent to contact to post-contact behavior—without requiring real-to-sim translation or object re-modeling. By eliminating the need to reconstruct physical objects in simulation, we reduce system setup complexity while enabling high-fidelity, real-world demonstration collection.}

\section{System Overview}
Our system offers an integrated workflow for controlling robot movements with both precision and safety, leveraging low-cost hardware (under \$1,000) and diverse input devices. This workflow comprises two complementary modules:

\begin{enumerate}
    \item \textbf{Teleoperation Module}: Includes a teleoperation pipeline for robust motion retargeting.
    \item \textbf{Preview-Based Module}: Allows operators to check intended movements in a virtual setting before execution.
\end{enumerate}

As illustrated in Fig.~\ref{fig: pipeline}, the \textit{teleoperation module} (Section~\ref{sec: preview}) offers a safe and intuitive pipeline that proceeds in three stages:
\begin{enumerate}
    \item Wrist pose estimation via IMU-based tracking (Section~\ref{sec: Wrist Pose Estimation}),
    \item Hand pose estimation using a motion capture glove (Section~\ref{sec: Hand-Joint-Retargeting}),
    \item Non-collision retargeting to ensure safe joint configurations (Section~\ref{sec: Non-Collision Retargeting}).
\end{enumerate}

Together, these components collectively enable an efficient mapping from human motion to robot commands. Moreover, by adopting SMPL-X (Section~\ref{sec: SMPL-X as a standard}) as a standardized representation of the human body, our pipeline can seamlessly integrate data from any input device conforming to this format.

Additionally, as illustrated in Fig.~\ref{fig: pipeline}, the \textit{preview-based module} (Section~\ref{sec: preview}) then provides a safe and intuitive pipeline that proceeds in three stages:
\begin{enumerate}
    \item The preview robot is aligned with the physical robot using AprilTag markers~\cite{olson2011apriltag} (Section~\ref{sec: align}).
    \item User commands are visualized on the aligned preview robot (Section~\ref{sec: previewvis}).
    \item Once the I/O state transitions\deleted{ from active to inactive}, the final preview pose is captured and converted into an optimal trajectory for actual execution (Section~\ref{sec: iocontrol}).
\end{enumerate}

This preview-then-execute scheme reduces errors and protects the robot and its surroundings by allowing users to visualize and refine their actions beforehand. Thus, it substantially improves efficiency and safety.

\section{Teleoperation Pipeline\label{sec:teleop}}
\subsection{Wrist Pose Estimation}
\label{sec: Wrist Pose Estimation}

Our system tracks the human operator's wrist motion to enable intuitive robot teleoperation. The input consists of raw IMU sensor data from sensors attached to the operator's arm, while the output is the 6-DoF wrist pose (position and orientation) in world coordinates. This wrist pose serves as the primary control signal for the robot's end-effector, allowing natural mapping between human arm movements and robot motion.

\subsubsection{SMPL-X Standard}
\label{sec: SMPL-X as a standard}
We adopt the SMPL-X standard \cite{pavlakos2019expressive} as our standard representation of the human body. The SMPL-X model represents a kinematic tree with standardized joint coordinate systems, where each joint's pose describes a rotation and translation relative to its parent frame. This hierarchical structure enables consistent pose representation across different motion capture devices and simplifies the integration with our TelePreview system.

\subsubsection{IMU-based Tracking System}

IMU-based motion capture offers robust tracking regardless of visual conditions and occlusions. Following the SMPL-X convention, we define the world frame system at the midpoint between the feet. All positions and orientations in this paper are expressed in this coordinate system unless otherwise specified. After obtaining the SMPL-X model, we compute the wrist position $\mathbf{p}_w \in \mathbb{R}^3$ and orientation $\mathbf{R}_w \in SO(3)$ relative to this world frame.

Based on the wrist parameters, we map the user's wrist pose to the end-effector pose through a transformation function $\mathcal{T}$. Specifically, we use the relative wrist position to control the end-effector position and the absolute wrist orientation to control the end-effector orientation:
\begin{equation}
\mathbf{p}_e(t) = \mathbf{p}_e(0) + (\mathbf{p}_w(t) - \mathbf{p}_w(0)), \quad
\mathbf{R}_e(t) = \mathbf{R}_w(t),
\end{equation}
where $\mathbf{p}_e(t)$ and $\mathbf{R}_e(t)$ represent the end-effector position and orientation at time $t$, and $\mathbf{p}_w(t)$ and $\mathbf{R}_w(t)$ represent the user's wrist position and orientation at time $t$.

\subsection{Hand Pose Estimation}
\label{sec: Hand-Joint-Retargeting}




Mocap gloves offer direct joint angle measurements $\{q_i\}_{i=1}^{27}$ using flex sensors. Our joint-to-joint mapping approach provides a generalizable solution for retargeting high-dimensional human hand motion data to lower-dimensional robotic systems. This approach is particularly effective when the input space (e.g., the human hand with 27 DoF) has higher dimensionality than the target space (e.g., the robotic hand with 16 DoF), as it allows selective mapping of the most intuitive and task-relevant degrees of freedom. Through a manual selection of corresponding joints based on human intuition and task requirements, followed by range alignment between input and output spaces, we maintain natural and predictable motion correspondence.

We propose a direct joint-to-joint mapping function $\mathcal{M}: \mathbb{R}^{27} \rightarrow \mathbb{R}^{16}$ that transforms mocap glove readings to LEAP hand \cite{shaw2023leaphand} joint configurations. Let $q_g \in \mathbb{R}^{27}$ denote the glove joint values and $q_r \in \mathbb{R}^{16}$ denote the robot joint values. For each robot joint $i$, we define a mapping:
\begin{equation}
q_r^{i} = f_i(q_g^{k_i}), \quad i \in {1,\ldots,16},
\end{equation}
where $k_i$ is the corresponding glove joint index for robot joint $i$ which is manually picked, and $f_i$ is a linear transformation:
\begin{equation}
f_i(x) = s_i(x - b_i)r_i.
\end{equation}
Here, $s_i \in \mathbb{R}^+$ is a scaling factor that normalizes the joint ranges, $b_i \in \mathbb{R}$ is a bias term that aligns the neutral positions, and $r_i \in$ \{-1,1\} is a direction indicator that ensures consistent joint rotations. This mapping guarantees:
\begin{equation}
\max q_g^{k^i} \Leftrightarrow \max q_r^i, \quad \min q_g^{k^i} \Leftrightarrow \min q_r^i,
\end{equation}
Where $q_g$ and $q_r$ represent the joint angles of the glove and the Leap Hand, respectively, and max and min denote their upper and lower joint limits.

\begin{figure}[htbp]
\centering
\begin{subfigure}{\linewidth}
\centering
\includegraphics[width=0.8\linewidth]{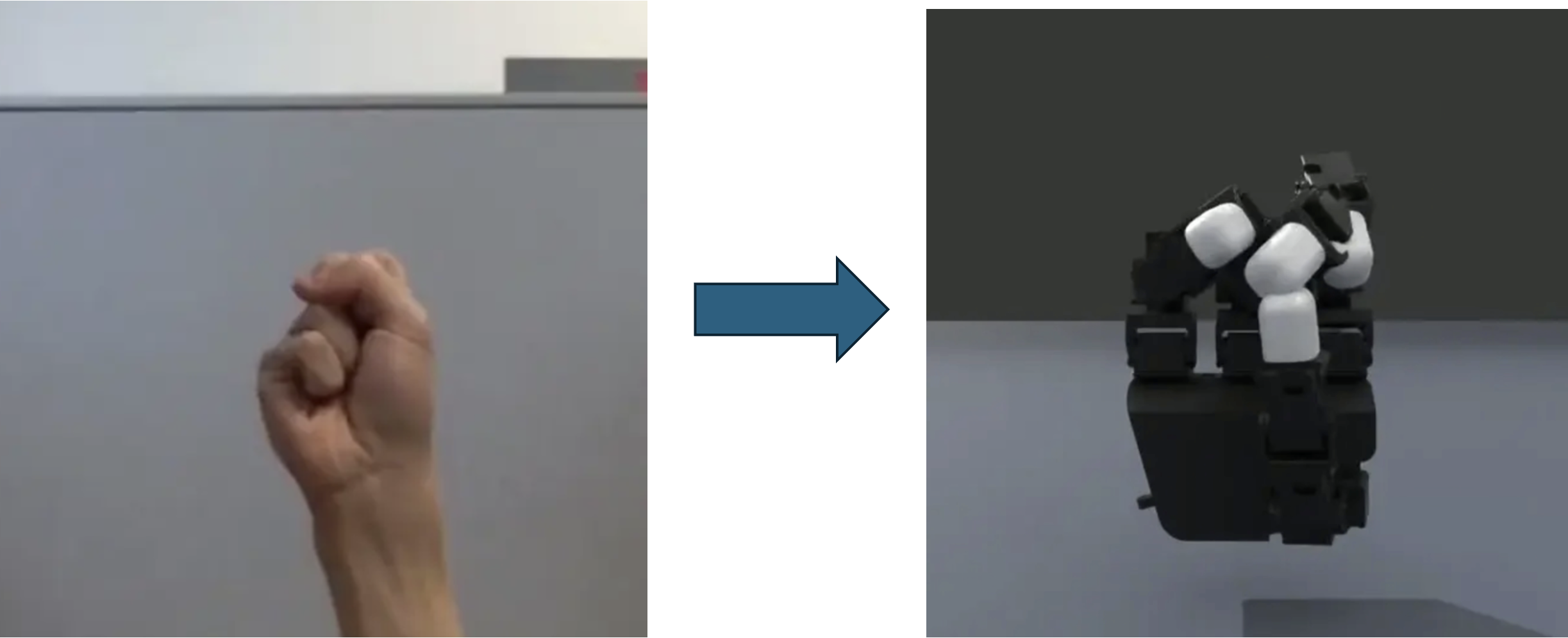}
\caption{Current Retargeting Method \cite{qin2023anyteleop} Causes Collision.}
\vspace{6px}
\label{fig:dex-retargeting1}
\end{subfigure}
\begin{subfigure}{\linewidth}
\centering
\includegraphics[width=0.8\linewidth]{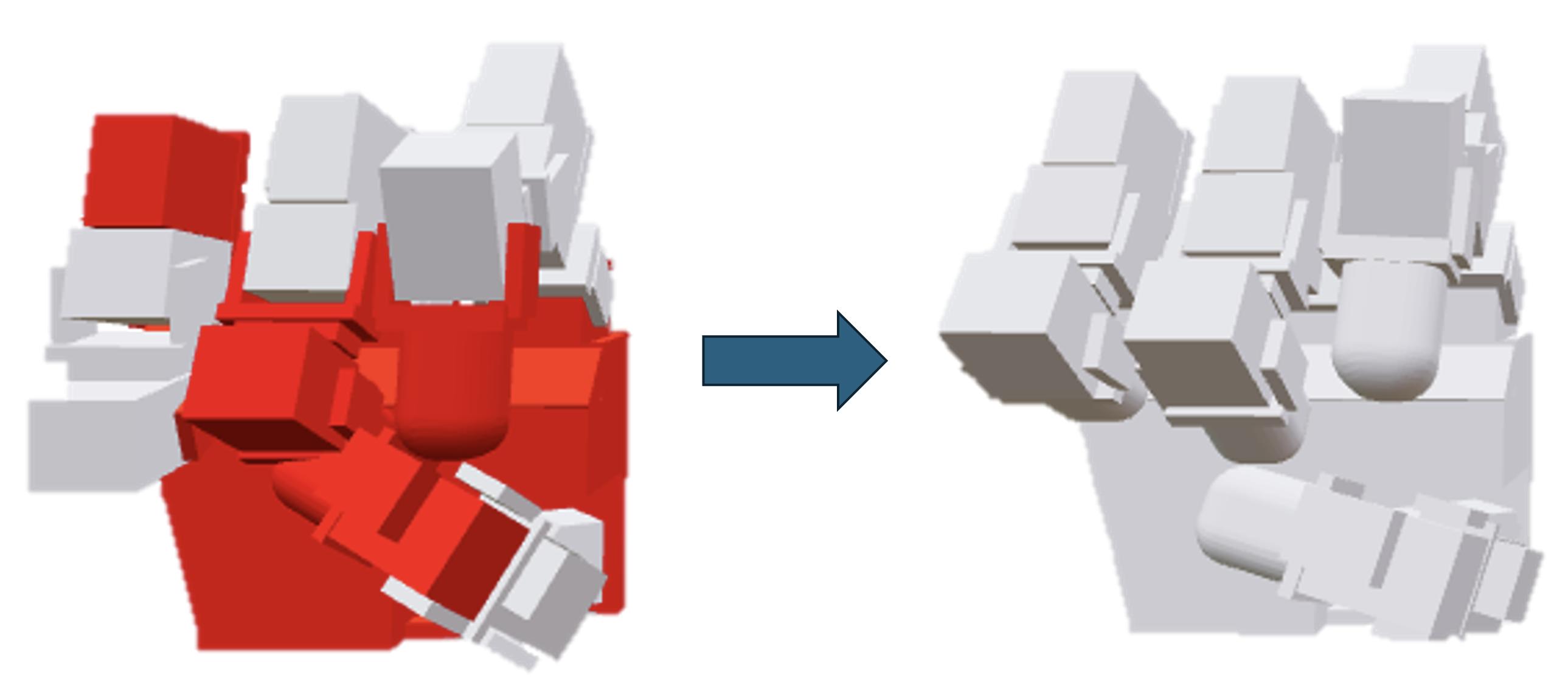}
\caption{Our Self-collision Avoidance Retargeting.}
\label{fig:collision-avoidance}
\end{subfigure}
\vspace{-6px}
\caption{\textbf{Comparison of Hand Configuration Retargeting Methods: }(a) shows the direct mapping between human and robot hands leading to self-collision; (b) demonstrates our collision-aware retargeting approach that maintains safe configurations.}
\vspace{-8px}
\label{fig:dex-retargeting}
\end{figure}

\subsection{Non-Collision Retargeting Network}
\label{sec: Non-Collision Retargeting}


Direct mapping from human hand postures to robotic hands using existing retargeting approaches like \cite{qin2023anyteleop} can result in self-collisions, as demonstrated in Fig.~\ref{fig:dex-retargeting1}. While traditional retargeting methods can achieve real-time performance for simpler end-effectors, they rely on optimization-based approaches that become computationally intensive for high-DoF configurations, \added{highlighting the importance of collision-aware methods such as}~\cite{sivakumar2022robotic}.

To address these challenges, we develop a learning-based framework with two key components:
\begin{enumerate}
    \item A Self-Collision Prediction Network (CPN) that takes robot joint configurations $q \in \mathbb{R}^n$ as input and outputs a binary collision probability vector $p \in \mathbb{R}^m$ indicating collision likelihood for each link.

    \item A Configuration Correction Network (CCN) that maps collision-prone joint configurations to their nearest collision-free counterparts. Specifically, it transforms an invalid robot configuration $q_{invalid} \in \mathbb{R}^n$ to a valid configuration $q_{valid} \in \mathbb{R}^n$ that minimizes both collision risk and deviation from the original pose.
\end{enumerate}

As demonstrated in Fig.~\ref{fig:collision-avoidance}, our method successfully transforms problematic configurations into stable, safe positions while maintaining efficient real-time performance (about 60Hz). \added{The implementation detail can be found in the Appendix.}

\section{Preview Pipeline\label{sec: preview}}
\subsection{Why Preview System Assistance?}
Traditional teleoperation approaches face significant challenges in data quality when collecting demonstrations for imitation learning. During complex manipulation tasks, users frequently make exploratory movements that contaminate the demonstration data with sub-optimal trajectories. \deleted{These repeated probing actions to locate the socket edge and refine the approach angle before successful insertion introduced significant noise into the collected datasets. }Such noise is particularly problematic for training imitation learning models, which benefit most from clean, purposeful demonstrations rather than trajectories cluttered with exploratory movements. \added{Recent research has highlighted how errors or exploratory movements in teleoperation can decrease the generalization ability of the learned policies and negatively impact performance on unseen tasks.}\cite{mandlekar2021matters, zhang2018deep, tangkaratt2020robust, hussein2023detecting, zheng2023extraneousness}

To address these limitations and provide a more efficient workflow, we propose incorporating a dedicated \emph{preview} feature that allows users to refine their actions before physical execution. By separating exploration from the final command, our approach reduces sub-optimal motions that would otherwise degrade data quality.

\subsection{Technical Implementation of Preview System}
\label{sec: preview_implementation}
We designed our preview feature to integrate seamlessly with the teleoperation workflow. It requires three critical elements: (1) precise spatial calibration to maintain alignment between the virtual preview and the physical robot, (2) realistic rendering to blend the preview with camera feeds, and (3) state management to handle transitions between preview and execution phases. The following subsections detail each of these technical components, highlighting how they collectively enable users to view and adjust planned motions prior to committing them to the physical robot.

\subsubsection{Spatial Alignment Using AprilTag}
\label{sec: align}

To maintain a precise spatial alignment between the preview and the physical robot in the camera viewpoints, we implement a robust calibration system using AprilTags~\cite{olson2011apriltag}. Our approach involves two sequential steps: (1) a hand-eye calibration that establishes the transformation between the robot's base frame and the fixed camera frame, and (2) AprilTag pose detection that enables real-time tracking of the robot's position and orientation relative to each camera view. \deleted{This chain of transformations ensures consistent preview visualization across different viewpoints.}

\begin{figure}[htbp]
\centering
\vspace{-8px}
\includegraphics[width=0.8 \linewidth]{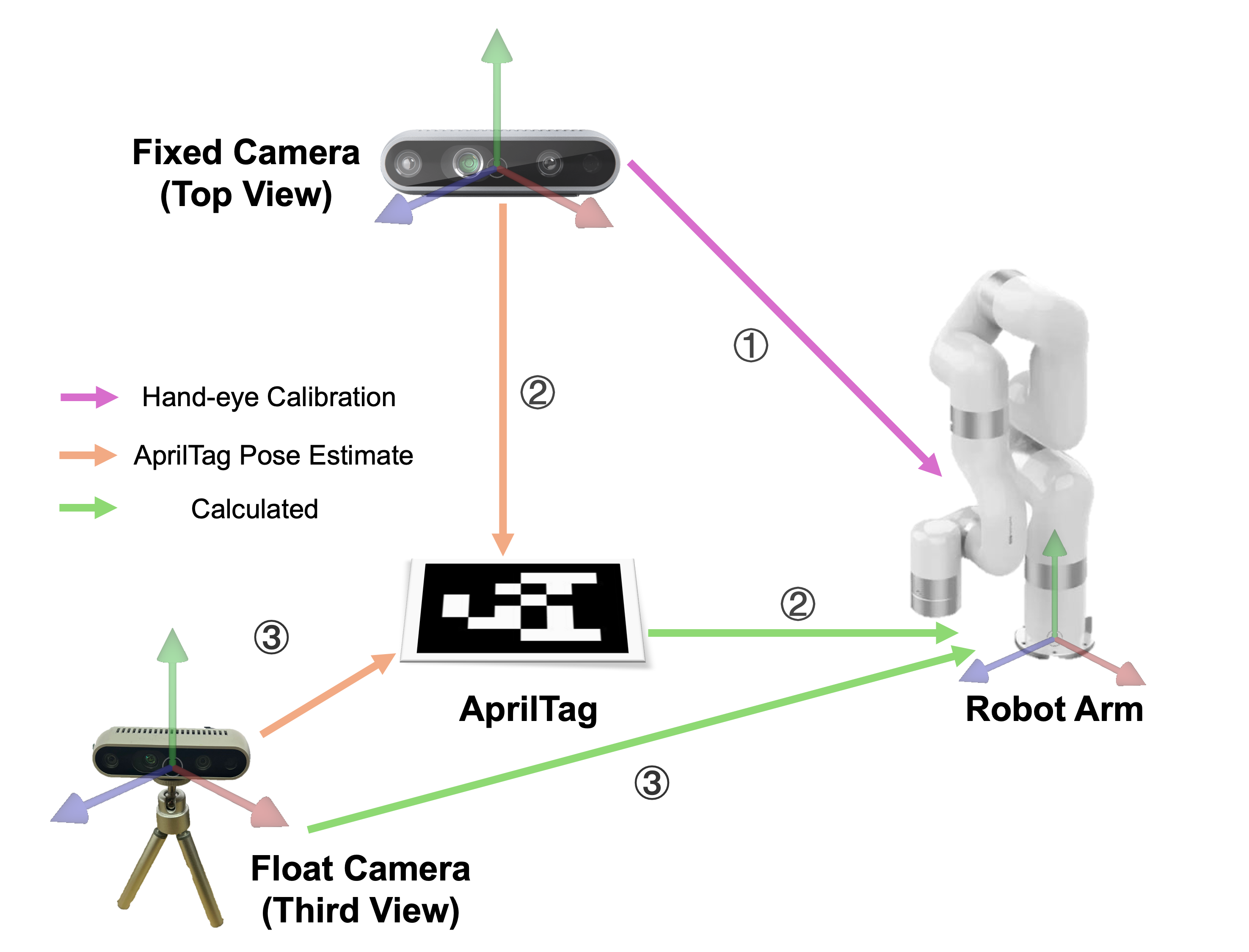}
\caption{\textbf{Our Transformation Relationship:} The number in the circle denotes the order of transformation acquisition.}
\vspace{-0.3cm}
\label{fig: relationship}
\end{figure}

As shown in Fig.~\ref{fig: relationship}, our calibration procedure involves multiple sequential transformations. We begin with hand-eye calibration of the fixed top-view camera, \deleted{(1),} followed by AprilTag pose detection from this camera's perspective. \deleted{(2)} \added{These steps allow us to compute the transformation from the AprilTag to the robot base frame.} For the floating third-view camera, we detect the AprilTag pose \deleted{from its viewpoint (3), which then allows us to establish its position relative to the robot base frame through the standard AprilTag reference} \added{and use it to determine its position relative to the robot base frame.} \deleted{(3)} \deleted{This chain of transformations ensures consistent preview visualization regardless of camera movement or viewpoint changes, making our system robust for dynamic viewing scenarios.} \added{This calibration ensures consistent visualization across dynamic or multi-view setups.}

\subsubsection{Visual Preview Visualization}
\label{sec: previewvis}
\begin{figure}[htbp]
\centering
\begin{subfigure}{0.5\linewidth}
   \centering
   \includegraphics[width=0.7\linewidth]{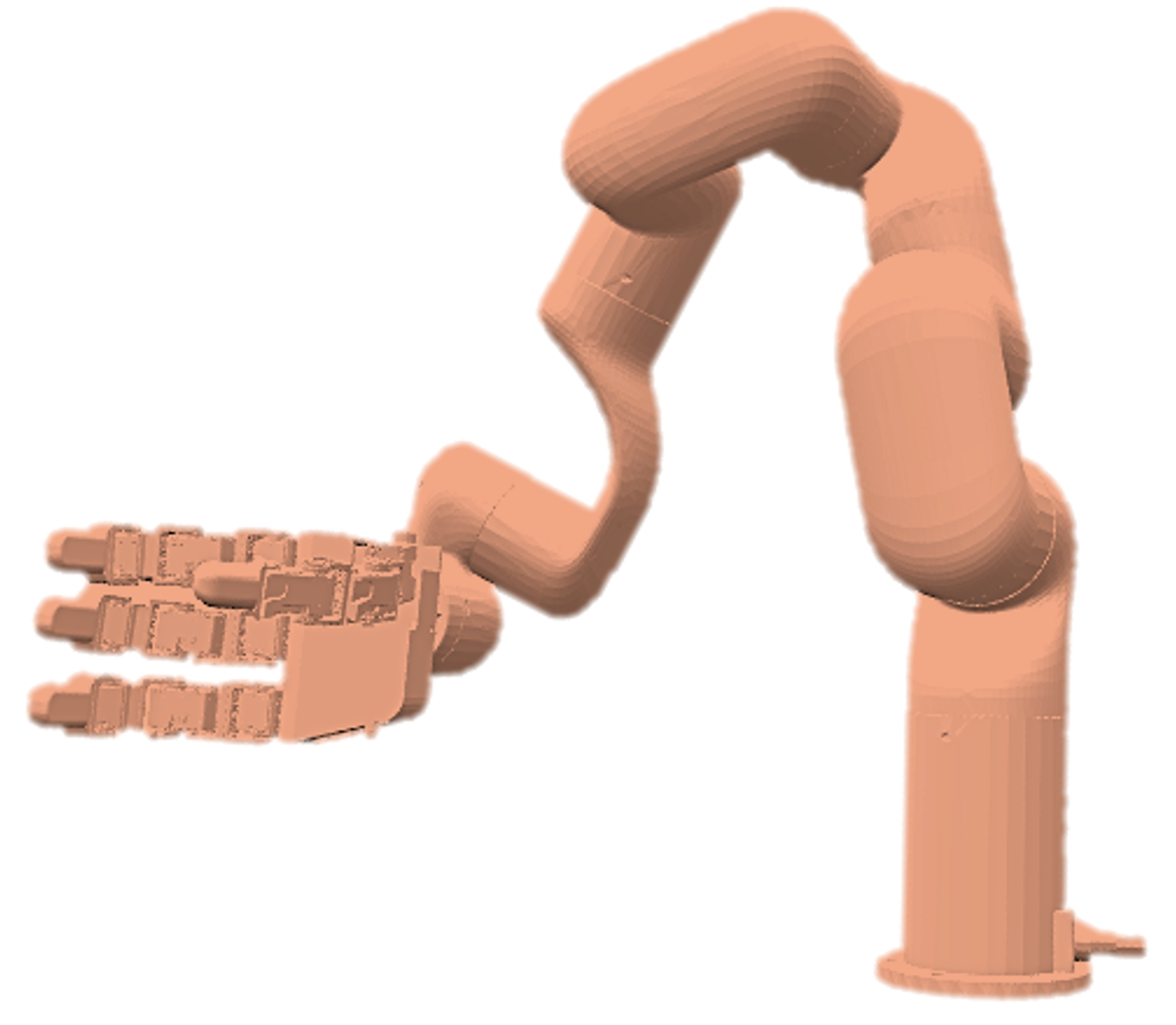}
   \caption{Preview Rendering}
   \label{fig: render}
\end{subfigure}
\\ \vspace{0.3cm}
\begin{subfigure}{0.4\linewidth}
   \centering
   \includegraphics[width=\linewidth]{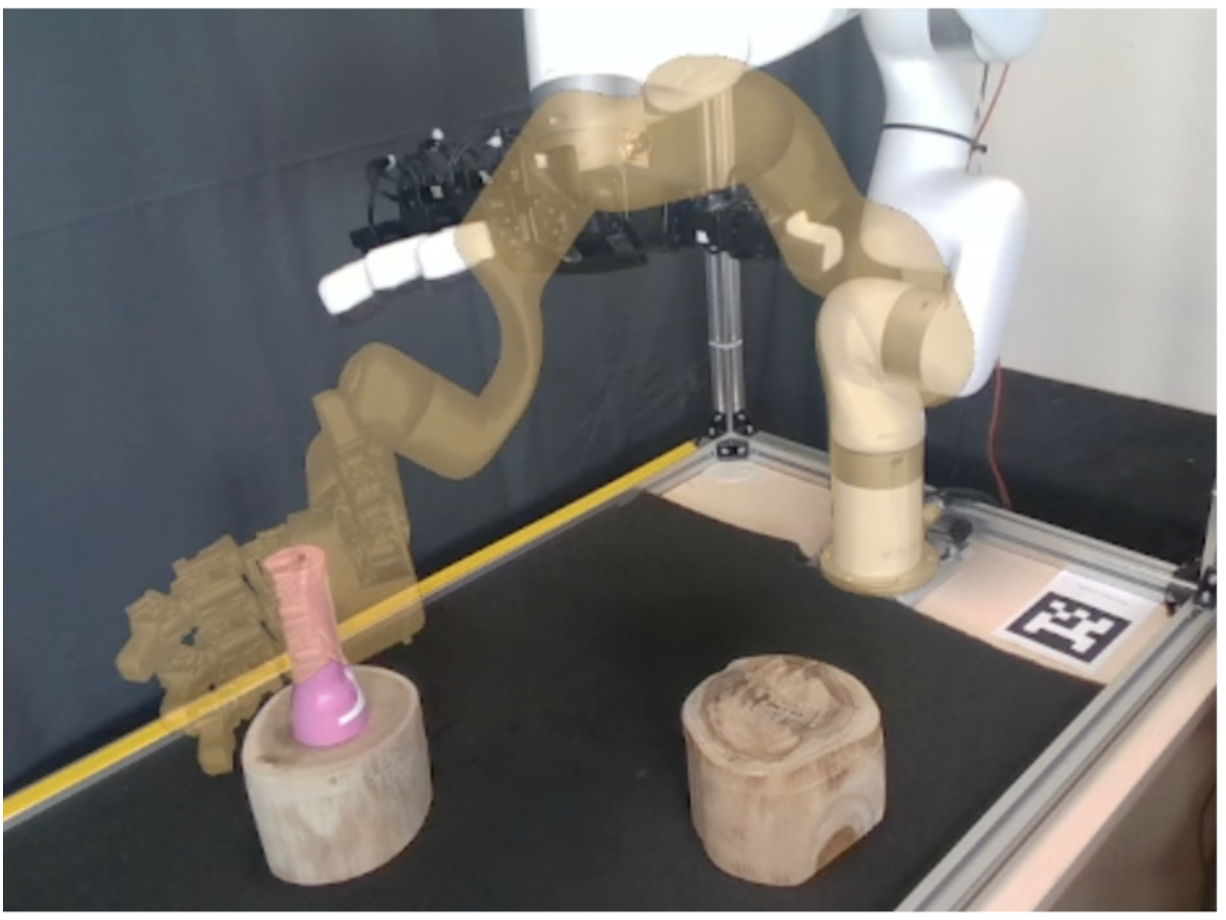}
   \caption{Third-person View}
   \label{fig: multi-view1}
\end{subfigure}
\begin{subfigure}{0.4\linewidth}
   \centering
   \includegraphics[width=\linewidth]{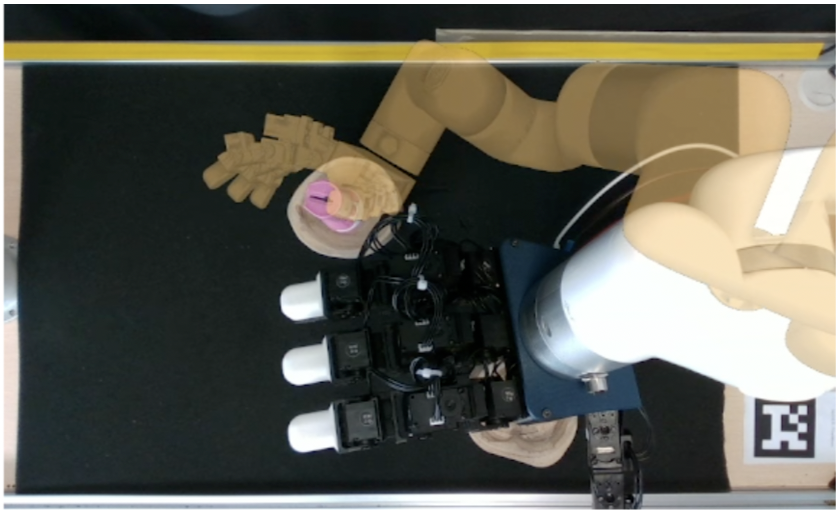}
   \caption{Top-down View}
   \label{fig: multi-view2}
\end{subfigure}
\caption{Rendering and Multi-view Visualization System.}
\label{fig: visualization}
\vspace{-0.5cm}
\end{figure}

Building upon this spatial alignment system, we implement the visual component of our preview system. As shown in Fig.~\ref{fig: render}, we use the pyrender library to render the virtual robot through the float camera. The preview image is then composited with the real camera feed using alpha blending, creating an overlay where the virtual robot appears aligned with the physical environment. This requires the calibration of virtual camera parameters to match the physical cameras, ensuring the preview accurately reflects the robot's intended configuration. \deleted{The resulting overlay provides users with an intuitive preview of planned motions.} \added{The preview offers real-time visual guidance before execution.} Our system supports multiple viewpoints \deleted{to enhance spatial awareness during complex manipulation tasks.} \added{to aid spatial reasoning and trajectory checking.} As shown in Fig.~\ref{fig: multi-view1} and Fig.~\ref{fig: multi-view2}, we deploy third-person and top-down cameras, \added{rendered and shown simultaneously on an external monitor. The system does not require a VR headset. For instance, during manipulation tasks, users can simultaneously monitor the robot's vertical position from the third-person view while checking its planar alignment from the top-down view.} \deleted{allowing users to verify planned movements from complementary perspectives.}

\subsubsection{IO Control for Preview Visualization}
\label{sec: iocontrol}

\begin{figure}[htbp]
\centering
\vspace{-5px}
\includegraphics[width=0.7\linewidth]{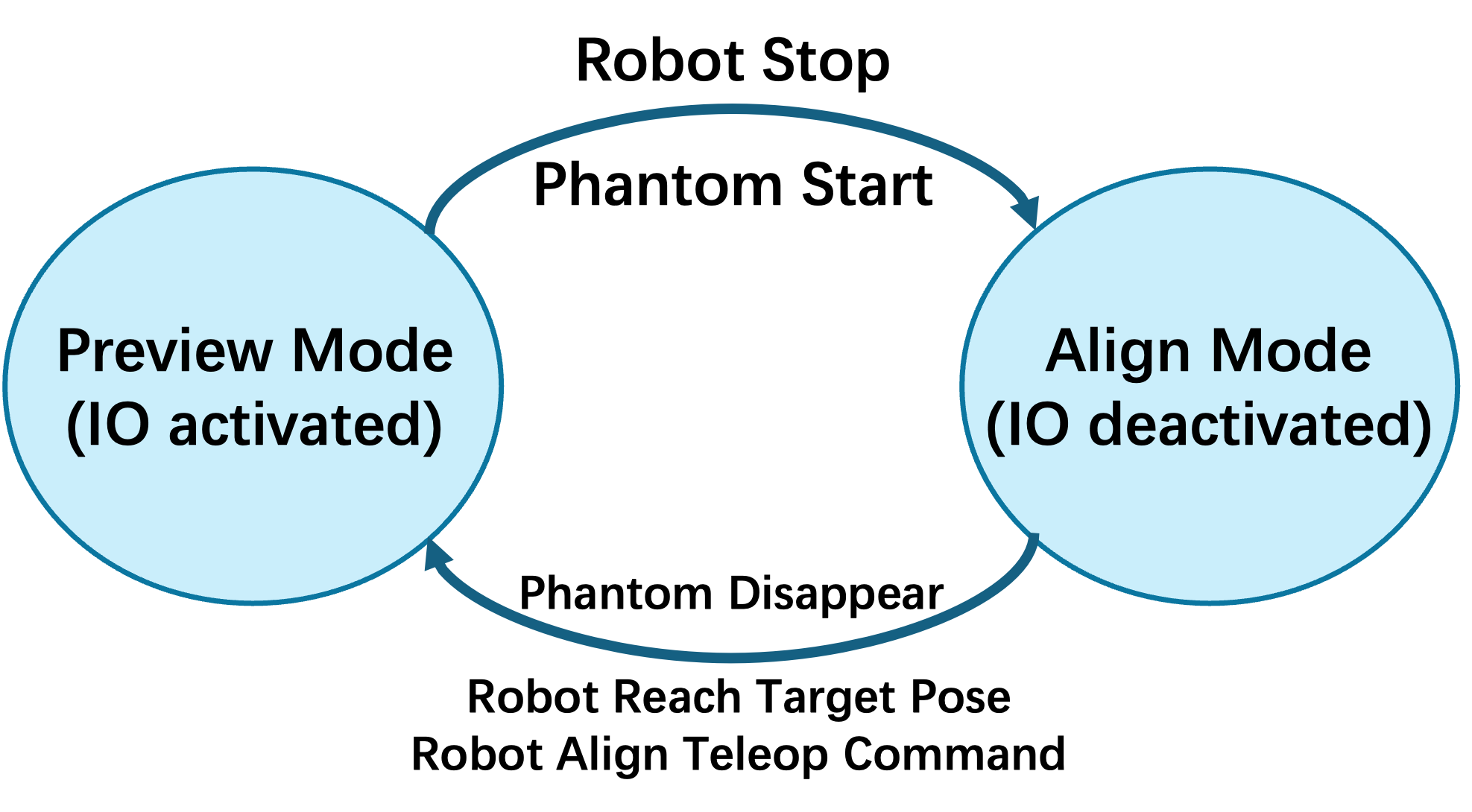}
\caption{State Transition of Preview Control}
\vspace{-0.5cm}
\label{fig: dfa}
\end{figure}

Our system uses a foot pedal as an intuitive control interface for state transitions. As shown in Fig.~\ref{fig: dfa}, when the IO is activated, the physical robot stops while the preview appears and responds to user commands for motion preview (Preview Mode). Upon IO deactivation, the preview disappears, and the system extracts its final configuration as the target pose to align, bypassing any exploratory movements made during the preview (Align Mode). The mplib motion planning library then processes this target to generate an optimal trajectory for execution. \added{During this motion, the user does not need to remain static—the robot executes the planned motion autonomously. Once the target pose is reached, the system enters a continuous teleoperation mode in which the robot follows the user's live movements in real time until the next preview is initiated.} \deleted{After reaching the target pose, the robot automatically resumes real-time teleoperation until the next IO is activated.} This workflow ensures that users execute only refined, intentional movements, eliminating exploratory motions from demonstration data.

\begin{table*}[ht!]
\vspace{0.5cm}
\centering
\begin{tabular}{lcccc}
    \toprule
    \textbf{Task} & \textbf{TelePreview} & \textbf{Open Teach \cite{iyer2024open}} & \textbf{AnyTeleop\cite{qin2023anyteleop}} & \textbf{Telekinesis\cite{sivakumar2022robotic}} \\
    \midrule
    Pick \& Place       & \textbf{1.0} & 0.8 & 
    \textbf{1.0} & 0.9 \\
    Hang          & \textbf{0.9} & - & - & - \\
    Pour  & \textbf{1.0} & 0.8 & 0.7 & 0.7 \\
    Box Rotation            & \textbf{1.0} & - & 0.6 & 0.6 \\
    Cup Stacking        & \textbf{1.0} & - & 0.7 & 0.3 \\
    \bottomrule
\end{tabular}
\caption{\textbf{Real Robot Teleoperation Results }\deleted{We use the success rates the baseline methods report in their papers, marking}\added{We mark} tasks they did not attempt or do not support with ``-''. Success rates for Open Teach \cite{iyer2024open} reflect expert performance.}
\label{result}
\vspace{-5px}
\end{table*}

\begin{table*}[ht!]
\centering
\begin{tabular}{lcccccc}
    \toprule
    \multirow{2.5}{*}{\textbf{Task}}            & \multicolumn{3}{c}{\textbf{Success Rate}} & \multicolumn{3}{c}{\textbf{Average Success Execution Time (s)}} \\
    \cmidrule(lr){2-4} \cmidrule(lr){5-7}
                             & \textbf{w/o preview} & \textbf{w/ preview} & \textbf{Difference $\uparrow$} & \textbf{w/o preview} & \textbf{w/ preview} & \textbf{Difference $\downarrow$} \\
    \midrule
    Pick \& Place   & 0.6   & 1.0   & \textbf{+0.4}   & 23.56\added{$\pm4.65$}   & 13.55\added{$\pm3.17$}   & \textbf{-10.01}    \\
    Hang            & 0.6   & 1.0   & \textbf{+0.4}   & 29.30\added{$\pm8.18$}   & 30.83\added{$\pm3.65$}   & +1.53     \\
    Pour            & 0.9   & 0.8   & -0.1   & 43.20\added{$\pm6.73$}   & 36.13\added{$\pm5.86$}   & \textbf{-7.07}     \\
    Box Rotation    & 0.6   & 0.8   & \textbf{+0.2}   & 30.47\added{$\pm14.08$}   & 19.12\added{$\pm3.32$}   & \textbf{-11.35}    \\
    Cup Stacking  & 0.5   & 1.0   & \textbf{+0.5}   & 31.54\added{$\pm6.04$}   & 18.91\added{$\pm1.29$}   & \textbf{-12.63}    \\
    \bottomrule
\end{tabular}
\caption{Effect of Preview Assistance on New User Performance.}
\vspace{-15px}
\label{improvement}
\end{table*}

\section{Experiments}
Our experiments aim to address the following questions:

Q1: How effective is TelePreview vs. Baselines?

Q2: How beneficial is TelePreview for new users?

Q3: How adaptable is TelePreview for new hardware?

\subsection{Experiment Setup}

TelePreview integrates several key components. We capture the user's movements via an inertial motion capture suit and a pair of strain-gauge-based mocap gloves, providing high-fidelity joint angle data for both body posture and delicate hand movements. To visualize robot action before execution, we employ a RealSense RGB-D camera and render the preview either on a standard 2D display or within a VR headset (e.g., Meta Quest 3). For the physical hardware, we use a UFactory xArm-6 robotic arm outfitted with a Leap hand \cite{shaw2023leaphand}, utilizing AprilTag \cite{olson2011apriltag} to align the initial poses of the preview and the real robot.

Following the methodology outlined in \cite{sivakumar2022robotic, qin2023anyteleop}, users attempted each task 10 times. We reference the baseline success rates from their papers.

\subsection{Task Descriptions}
\label{sec: tasks}
We test TelePreview on five real-world tasks\deleted{, each highlighting different manipulation challenges}. \added{These tasks were carefully selected to cover a diverse range of manipulation skills, including grasping, rotation, alignment, and insertion, and are representative of common benchmarks used in teleoperation research}:

\begin{itemize}
   \item \textbf{Pick \& Place:} Grasp an object and place it at precise locations.
   \item \textbf{Pour:} Tilt and rotate a cup to pour the beans into another container.
   \item \textbf{Hang:} Hang a spoon on a peg, requiring both fine positioning and subtle wrist rotations.
   \item \textbf{Box Rotation:} Rotate a box to change its orientation.
   \item \textbf{Cup Stacking:} Stack a cup onto another cup.
\end{itemize}

\subsection{Evaluation Metrics}
We measure two key metrics for each task:
\begin{itemize}
    \item \textbf{Success Rate:} We define the success rate as the proportion of times the operator completes the task within the given constraints.\deleted{ For example, if the user attempts a cup-hanging task \(X\) times and successfully hangs the cup on the rack \(Y\) times, the success rate is \(\frac{Y}{X}\).}
    \item \textbf{Execution Time:} The total duration from task initiation to successful completion (i.e., the length of the recorded demo episode).
\end{itemize}

\subsection{Performance Comparison with Baselines (Q1)}
\label{sec:performance_compare}

\added{Due to frequent self-collision failures encountered during local deployment of vision-based methods, we report results from their original publications and discuss reproduction experiences in the Appendix.} As summarized in Table~\ref{result}, TelePreview achieves higher success rates than all baselines. While Telekinesis, AnyTeleop, and Open Teach handle simpler, short-horizon tasks, they often struggle with complex manipulations requiring subtle wrist rotations and fine-grained positioning (e.g., the Hang task). TelePreview, by contrast, captures these nuances more effectively, reflecting stronger overall performance and versatility. \deleted{These findings demonstrate TelePreview's ability to address a broad spectrum of real-world challenges, from basic pick-and-place operations to more intricate tasks involving delicate movements.}

\subsection{Effect of Preview for Users (Q2)}
\label{sec:effect_preview}
To investigate \textbf{Q2}—how the preview feature benefits new (non-expert) users—we evaluate both \textbf{success rate} and \textbf{average execution time} in two conditions: (1) with the preview enabled and (2) with the preview disabled. \added{Our study involved 10 participants in total, including 5 new users with no prior experience in robotic teleoperation and 5 experienced users (3 of whom are authors). This section focuses on the new user group.}

\added{Participants were given consistent task instructions and evaluated under a standardized protocol, including predefined success criteria. Task order was counterbalanced across participants to mitigate ordering effects.}

The same five tasks from Section~\ref{sec: tasks} are attempted by new users, with each task repeated 10 times. \added{We report both mean and standard deviation across trials for each task. In addition to objective performance, we collected subjective workload ratings using a 5-point Likert scale (using the same evaluation metrics in }\cite{li2025trainrobotsimpactdemonstration}\added{) and recorded the practice time required for participants to reach task readiness under different input modalities. Detailed results and analysis are provided in the Appendix.}

As shown in Table~\ref{improvement}, enabling the preview boosts user performance. Success rates increase across all tasks, with especially notable gains in challenging manipulations such as \emph{cup stacking} (+0.5) and \emph{pick \& place} (+0.4). Moreover, average execution times drop by up to 12.63 seconds (cup stacking), demonstrating not only improved accuracy but also greater efficiency. \added{The standard deviations across trials are also reduced, suggesting more consistent user behavior.}

By visualizing a virtual ``preview'' of the robot's intended motion and confirming the final pose prior to executing physical movements, users eliminate the need for exploratory motions in the actual workspace. This design ensures that recorded trajectories contain only intentional, task-oriented actions \deleted{without extraneous trial-and-error, providing cleaner, high-quality demonstrations for imitation learning}. \added{This results in higher-quality demonstrations that are free from unnecessary trial-and-error behaviors.} \deleted{As a result, TelePreview delivers both a more intuitive teleoperation experience and more consistent data for downstream training.}

\subsection{Deployment on Various Robots (Q3)}
\deleted{We deploy our TelePreview system on a Ufactory xArm equipped with both a LEAP hand and a parallel gripper. By updating only the kinematic parameters and transformation chain for each end-effector, TelePreview preserves its full functionality across diverse setups, as shown in the Figure. This successful deployment on multiple end-effector configurations illustrates the system's adaptability to new hardware. Its modular design allows users to seamlessly integrate TelePreview into different robotic platforms while retaining all key features - preview, multi-view visualization, and intuitive control interfaces.}

\added{We evaluate the generalizability of TelePreview across three end-effector types on a Ufactory xArm: a 16-DoF anthropomorphic LEAP hand, a parallel-jaw gripper, and a vacuum gripper (both 1-DoF). Our results (see Appendix) show that TelePreview yields the most significant performance gains with the LEAP hand, highlighting its effectiveness for high-DoF, fine-grained manipulation tasks. Improvements with the gripper and vacuum gripper are more modest but still demonstrate the system's utility across diverse hardware configurations. These findings support our claim that TelePreview generalizes well while providing the greatest benefit in dexterous settings.}

\section{Conclusion}
In this paper, we presented TelePreview, a low-cost teleoperation system featuring AR preview feedback designed for deployment across diverse robotic platforms. Our experiments demonstrate that TelePreview offers sufficient flexibility and safety to perform a variety of fine-grained tasks successfully. TelePreview provides immediate, robot-centric feedback that enables users to preview and refine their commands before physical execution.

While our results show promising advances in teleoperation interfaces, a key limitation is the visual ambiguity caused by occlusions between the preview robot and scene objects. \deleted{Future work could address this by incorporating in-depth information from RGB-D cameras to create more accurate spatial relationships between virtual and physical elements.} \added{This limitation stems from our use of alpha blending, which lacks depth‑aware occlusion reasoning and cannot distinguish whether the virtual arm should occlude real objects or vice versa, resulting in artifacts even on passthrough‑capable devices like the Meta Quest 3 without real‑time depth estimation. To mitigate this ambiguity, our approach employs multi‑view overlays (e.g., top‑down and side views), allowing operators to compare perspectives and correctly infer spatial ordering.}

\deleted{Addressing these visualization challenges could yield several essential benefits. Improved occlusion handling would reduce user uncertainty during complex manipulation tasks, leading to higher-quality demonstration data for imitation learning. Additionally, more accurate spatial rendering would create a more intuitive teleoperation experience, potentially reducing user training time and cognitive load.}

\added{Future work will explore incorporating depth estimation from RGB-D or passthrough feeds and using it to inform depth-aware rendering and spatial reasoning. These could reduce user uncertainty during manipulation, leading to cleaner demonstration data and a more intuitive teleoperation experience.} 

{\small
\bibliographystyle{IEEEtran}
\bibliography{references}

\begin{thebibliography}{10}
\providecommand{\url}[1]{#1}
\csname url@samestyle\endcsname
\providecommand{\newblock}{\relax}
\providecommand{\bibinfo}[2]{#2}
\providecommand{\BIBentrySTDinterwordspacing}{\spaceskip=0pt\relax}
\providecommand{\BIBentryALTinterwordstretchfactor}{4}
\providecommand{\BIBentryALTinterwordspacing}{\spaceskip=\fontdimen2\font plus
\BIBentryALTinterwordstretchfactor\fontdimen3\font minus \fontdimen4\font\relax}
\providecommand{\BIBforeignlanguage}[2]{{%
\expandafter\ifx\csname l@#1\endcsname\relax
\typeout{** WARNING: IEEEtran.bst: No hyphenation pattern has been}%
\typeout{** loaded for the language `#1'. Using the pattern for}%
\typeout{** the default language instead.}%
\else
\language=\csname l@#1\endcsname
\fi
#2}}
\providecommand{\BIBdecl}{\relax}
\BIBdecl

\bibitem{iyer2024open}
A.~Iyer, Z.~Peng, Y.~Dai, I.~Guzey, S.~Haldar, S.~Chintala, and L.~Pinto, ``Open teach: A versatile teleoperation system for robotic manipulation,'' \emph{arXiv preprint arXiv:2403.07870}, 2024.

\bibitem{fu2024mobile}
Z.~Fu, T.~Z. Zhao, and C.~Finn, ``Mobile aloha: Learning bimanual mobile manipulation with low-cost whole-body teleoperation,'' \emph{arXiv preprint arXiv:2401.02117}, 2024.

\bibitem{qin2023anyteleop}
Y.~Qin, W.~Yang, B.~Huang, K.~Van~Wyk, H.~Su, X.~Wang, Y.-W. Chao, and D.~Fox, ``Anyteleop: A general vision-based dexterous robot arm-hand teleoperation system,'' \emph{arXiv preprint arXiv:2307.04577}, 2023.

\bibitem{shaw2024bimanual}
K.~Shaw, Y.~Li, J.~Yang, M.~K. Srirama, R.~Liu, H.~Xiong, R.~Mendonca, and D.~Pathak, ``Bimanual dexterity for complex tasks,'' \emph{arXiv preprint arXiv:2411.13677}, 2024.

\bibitem{park2024using}
Y.~Park and P.~Agrawal, ``Using apple vision pro to train and control robots,'' 2024.

\bibitem{wang2024dexcap}
C.~Wang, H.~Shi, W.~Wang, R.~Zhang, L.~Fei-Fei, and C.~K. Liu, ``Dexcap: Scalable and portable mocap data collection system for dexterous manipulation,'' \emph{arXiv preprint arXiv:2403.07788}, 2024.

\bibitem{cheng2024open}
X.~Cheng, J.~Li, S.~Yang, G.~Yang, and X.~Wang, ``Open-television: Teleoperation with immersive active visual feedback,'' \emph{arXiv preprint arXiv:2407.01512}, 2024.

\bibitem{kennel2021manipulability}
F.~Kennel-Maushart, R.~Poranne, and S.~Coros, ``Manipulability optimization for multi-arm teleoperation,'' in \emph{ICRA 2021}.\hskip 1em plus 0.5em minus 0.4em\relax IEEE, 2021, pp. 3956--3962.

\bibitem{huang2025human}
Y.~Huang, D.~Fan, H.~Duan, D.~Yan, W.~Qi, J.~Sun, Q.~Liu, and P.~Wang, ``Human-like dexterous manipulation for anthropomorphic five-fingered hands: A review,'' \emph{Biomimetic Intelligence and Robotics}, p. 100212, 2025.

\bibitem{darvish2023teleoperation}
K.~Darvish, L.~Penco, J.~Ramos, R.~Cisneros, J.~Pratt, E.~Yoshida, S.~Ivaldi, and D.~Pucci, ``Teleoperation of humanoid robots: A survey,'' \emph{IEEE Transactions on Robotics}, vol.~39, no.~3, pp. 1706--1727, 2023.

\bibitem{lichiardopol2007survey}
S.~Lichiardopol, ``A survey on teleoperation,'' Technische Universiteit Eindhoven, Technical Report DCT-2007.155, 2007.

\bibitem{hokayem2006bilateral}
P.~F. Hokayem and M.~W. Spong, ``Bilateral teleoperation: An historical survey,'' \emph{Automatica}, vol.~42, no.~12, pp. 2035--2057, 2006.

\bibitem{katyal2014approaches}
K.~D. Katyal, C.~Y. Brown, S.~A. Hechtman, M.~P. Para, T.~G. McGee, K.~C. Wolfe, R.~J. Murphy, M.~D. Kutzer, E.~W. Tunstel, M.~P. McLoughlin \emph{et~al.}, ``Approaches to robotic teleoperation in a disaster scenario: From supervised autonomy to direct control,'' in \emph{2014 IEEE/RSJ International Conference on Intelligent Robots and Systems}.\hskip 1em plus 0.5em minus 0.4em\relax IEEE, 2014, pp. 1874--1881.

\bibitem{aronson2018eye}
R.~M. Aronson, T.~Santini, T.~C. K{\"u}bler, E.~Kasneci, S.~Srinivasa, and H.~Admoni, ``Eye-hand behavior in human-robot shared manipulation,'' in \emph{Proceedings of the 2018 ACM/IEEE International Conference on Human-Robot Interaction}, 2018, pp. 4--13.

\bibitem{scherzinger2023learning}
S.~Scherzinger, A.~Roennau, and R.~Dillmann, ``Learning human-inspired force strategies for robotic assembly,'' in \emph{2023 IEEE 19th International Conference on Automation Science and Engineering (CASE)}.\hskip 1em plus 0.5em minus 0.4em\relax IEEE, 2023, pp. 1--8.

\bibitem{micire2011design}
M.~Micire, M.~Desai, J.~L. Drury, E.~McCann, A.~Norton, K.~M. Tsui, and H.~A. Yanco, ``Design and validation of two-handed multi-touch tabletop controllers for robot teleoperation,'' in \emph{Proceedings of the 16th international conference on Intelligent user interfaces}, 2011, pp. 145--154.

\bibitem{ding2024bunny}
R.~Ding, Y.~Qin, J.~Zhu, C.~Jia, S.~Yang, R.~Yang, X.~Qi, and X.~Wang, ``Bunny-visionpro: Real-time bimanual dexterous teleoperation for imitation learning,'' \emph{arXiv preprint arXiv:2407.03162}, 2024.

\bibitem{9035064}
M.~Hirschmanner, C.~Tsiourti, T.~Patten, and M.~Vincze, ``Virtual reality teleoperation of a humanoid robot using markerless human upper body pose imitation,'' in \emph{2019 IEEE-RAS 19th International Conference on Humanoid Robots (Humanoids)}, 2019, pp. 259--265.

\bibitem{stanton2012teleoperation}
C.~Stanton, A.~Bogdanovych, and E.~Ratanasena, ``Teleoperation of a humanoid robot using full-body motion capture, example movements, and machine learning,'' in \emph{Proc. Australasian Conference on Robotics and Automation}, vol.~8, 2012, p.~51.

\bibitem{miller2004motion}
N.~Miller, O.~C. Jenkins, M.~Kallmann, and M.~J. Mataric, ``Motion capture from inertial sensing for untethered humanoid teleoperation,'' in \emph{4th IEEE/RAS International Conference on Humanoid Robots, 2004.}, vol.~2.\hskip 1em plus 0.5em minus 0.4em\relax IEEE, 2004, pp. 547--565.

\bibitem{liu2019high}
H.~Liu, Z.~Zhang, X.~Xie, Y.~Zhu, Y.~Liu, Y.~Wang, and S.-C. Zhu, ``High-fidelity grasping in virtual reality using a glove-based system,'' in \emph{ICRA 2019}.\hskip 1em plus 0.5em minus 0.4em\relax IEEE, 2019, pp. 5180--5186.

\bibitem{mosbach2022accelerating}
M.~Mosbach, K.~Moraw, and S.~Behnke, ``Accelerating interactive human-like manipulation learning with gpu-based simulation and high-quality demonstrations,'' in \emph{2022 IEEE-RAS 21st International Conference on Humanoid Robots (Humanoids)}.\hskip 1em plus 0.5em minus 0.4em\relax IEEE, 2022, pp. 435--441.

\bibitem{schwarz2021nimbro}
M.~Schwarz, C.~Lenz, A.~Rochow, M.~Schreiber, and S.~Behnke, ``Nimbro avatar: Interactive immersive telepresence with force-feedback telemanipulation,'' in \emph{2021 IEEE/RSJ International Conference on Intelligent Robots and Systems (IROS)}.\hskip 1em plus 0.5em minus 0.4em\relax IEEE, 2021, pp. 5312--5319.

\bibitem{audonnet2024telesim}
F.~P. Audonnet, J.~Grizou, A.~Hamilton, and G.~Aragon-Camarasa, ``Telesim: A modular and plug-and-play framework for robotic arm teleoperation using a digital twin,'' in \emph{ICRA 2024}.\hskip 1em plus 0.5em minus 0.4em\relax IEEE, 2024, pp. 17\,770--17\,777.

\bibitem{yang2024ace}
S.~Yang, M.~Liu, Y.~Qin, R.~Ding, J.~Li, X.~Cheng, R.~Yang, S.~Yi, and X.~Wang, ``Ace: A cross-platform visual-exoskeletons system for low-cost dexterous teleoperation,'' \emph{arXiv preprint arXiv:2408.11805}, 2024.

\bibitem{wu2023gello}
P.~Wu, Y.~Shentu, Z.~Yi, X.~Lin, and P.~Abbeel, ``Gello: A general, low-cost, and intuitive teleoperation framework for robot manipulators,'' 2023.

\bibitem{lawrence1993stability}
D.~A. Lawrence, ``Stability and transparency in bilateral teleoperation,'' \emph{IEEE transactions on robotics and automation}, vol.~9, no.~5, pp. 624--637, 1993.

\bibitem{ousaid2015stable}
A.~M. Ousaid, D.~S. Haliyo, S.~R{\'e}gnier, and V.~Hayward, ``A stable and transparent microscale force feedback teleoperation system,'' \emph{IEEE/ASME Transactions on Mechatronics}, vol.~20, no.~5, pp. 2593--2603, 2015.

\bibitem{zhao2017augmented}
Z.~Zhao, P.~Huang, Z.~Lu, and Z.~Liu, ``Augmented reality for enhancing tele-robotic system with force feedback,'' \emph{Robotics and Autonomous Systems}, vol.~96, pp. 93--101, 2017.

\bibitem{lee2018implementation}
D.~Lee and Y.~S. Park, ``Implementation of augmented teleoperation system based on robot operating system (ros),'' in \emph{2018 IEEE/RSJ International Conference on Intelligent Robots and Systems (IROS)}.\hskip 1em plus 0.5em minus 0.4em\relax IEEE, 2018, pp. 5497--5502.

\bibitem{milgram1995telerobotic}
P.~Milgram, A.~Rastogi, and J.~J. Grodski, ``Telerobotic control using augmented reality,'' in \emph{Proceedings 4th IEEE International Workshop on Robot and Human Communication}.\hskip 1em plus 0.5em minus 0.4em\relax IEEE, 1995, pp. 21--29.

\bibitem{sivakumar2022robotic}
A.~Sivakumar, K.~Shaw, and D.~Pathak, ``Robotic telekinesis: Learning a robotic hand imitator by watching humans on youtube,'' \emph{arXiv preprint arXiv:2202.10448}, 2022.

\bibitem{wang2024eve}
J.~Wang, C.-C. Chang, J.~Duan, D.~Fox, and R.~Krishna, ``Eve: Enabling anyone to train robots using augmented reality,'' in \emph{Proceedings of the 37th Annual ACM Symposium on User Interface Software and Technology}, 2024, pp. 1--13.

\bibitem{duan2023ar2}
J.~Duan, Y.~R. Wang, M.~Shridhar, D.~Fox, and R.~Krishna, ``Ar2-d2: Training a robot without a robot,'' \emph{arXiv preprint arXiv:2306.13818}, 2023.

\bibitem{olson2011apriltag}
E.~Olson, ``Apriltag: A robust and flexible visual fiducial system,'' in \emph{2011 IEEE international conference on robotics and automation}.\hskip 1em plus 0.5em minus 0.4em\relax IEEE, 2011, pp. 3400--3407.

\bibitem{pavlakos2019expressive}
G.~Pavlakos, V.~Choutas, N.~Ghorbani, T.~Bolkart, A.~A. Osman, D.~Tzionas, and M.~J. Black, ``Expressive body capture: 3d hands, face, and body from a single image,'' in \emph{CVPR}, 2019, pp. 10\,975--10\,985.

\bibitem{shaw2023leaphand}
K.~Shaw, A.~Agarwal, and D.~Pathak, ``Leap hand: Low-cost, efficient, and anthropomorphic hand for robot learning,'' \emph{Robotics: Science and Systems (RSS)}, 2023.

\bibitem{mandlekar2021matters}
A.~Mandlekar, D.~Xu, J.~Wong, S.~Nasiriany, C.~Wang, R.~Kulkarni, L.~Fei-Fei, S.~Savarese, Y.~Zhu, and R.~Mart{\'\i}n-Mart{\'\i}n, ``What matters in learning from offline human demonstrations for robot manipulation,'' \emph{arXiv preprint arXiv:2108.03298}, 2021.

\bibitem{zhang2018deep}
T.~Zhang, Z.~McCarthy, O.~Jow, D.~Lee, X.~Chen, K.~Goldberg, and P.~Abbeel, ``Deep imitation learning for complex manipulation tasks from virtual reality teleoperation,'' in \emph{ICRA 2018}.\hskip 1em plus 0.5em minus 0.4em\relax Ieee, 2018, pp. 5628--5635.

\bibitem{tangkaratt2020robust}
V.~Tangkaratt, N.~Charoenphakdee, and M.~Sugiyama, ``Robust imitation learning from noisy demonstrations,'' \emph{arXiv preprint arXiv:2010.10181}, 2020.

\bibitem{hussein2023detecting}
M.~Hussein and M.~Begum, ``Detecting incorrect visual demonstrations for improved policy learning,'' in \emph{Conference on Robot Learning}.\hskip 1em plus 0.5em minus 0.4em\relax PMLR, 2023, pp. 1817--1827.

\bibitem{zheng2023extraneousness}
R.~C. Zheng, K.~Hu, Z.~Yuan, B.~Chen, and H.~Xu, ``Extraneousness-aware imitation learning,'' in \emph{ICRA 2023}.\hskip 1em plus 0.5em minus 0.4em\relax IEEE, 2023, pp. 2973--2979.

\bibitem{li2025trainrobotsimpactdemonstration}
\BIBentryALTinterwordspacing
H.~Li, Y.~Cui, and D.~Sadigh, ``How to train your robots? the impact of demonstration modality on imitation learning,'' 2025. [Online]. Available: \url{https://arxiv.org/abs/2503.07017}
\BIBentrySTDinterwordspacing

\end{thebibliography}
}

\clearpage
\appendix
\subsection{Technical Implementation of Pos-to-Pos Non-Collision Module}

Due to differences in action space and limitations in the precision of the retargeting algorithm, the configuration of a dexterous robotic hand often generates invalid self-collision configurations. These invalid configurations not only lack operational utility but also risk system failure or hardware damage. To address this issue, we propose a method for mapping invalid configurations to their closest valid counterparts, enabling recovery from self-collision scenarios.

\begin{figure}[htbp]
  \centering
\includegraphics[width=\linewidth]{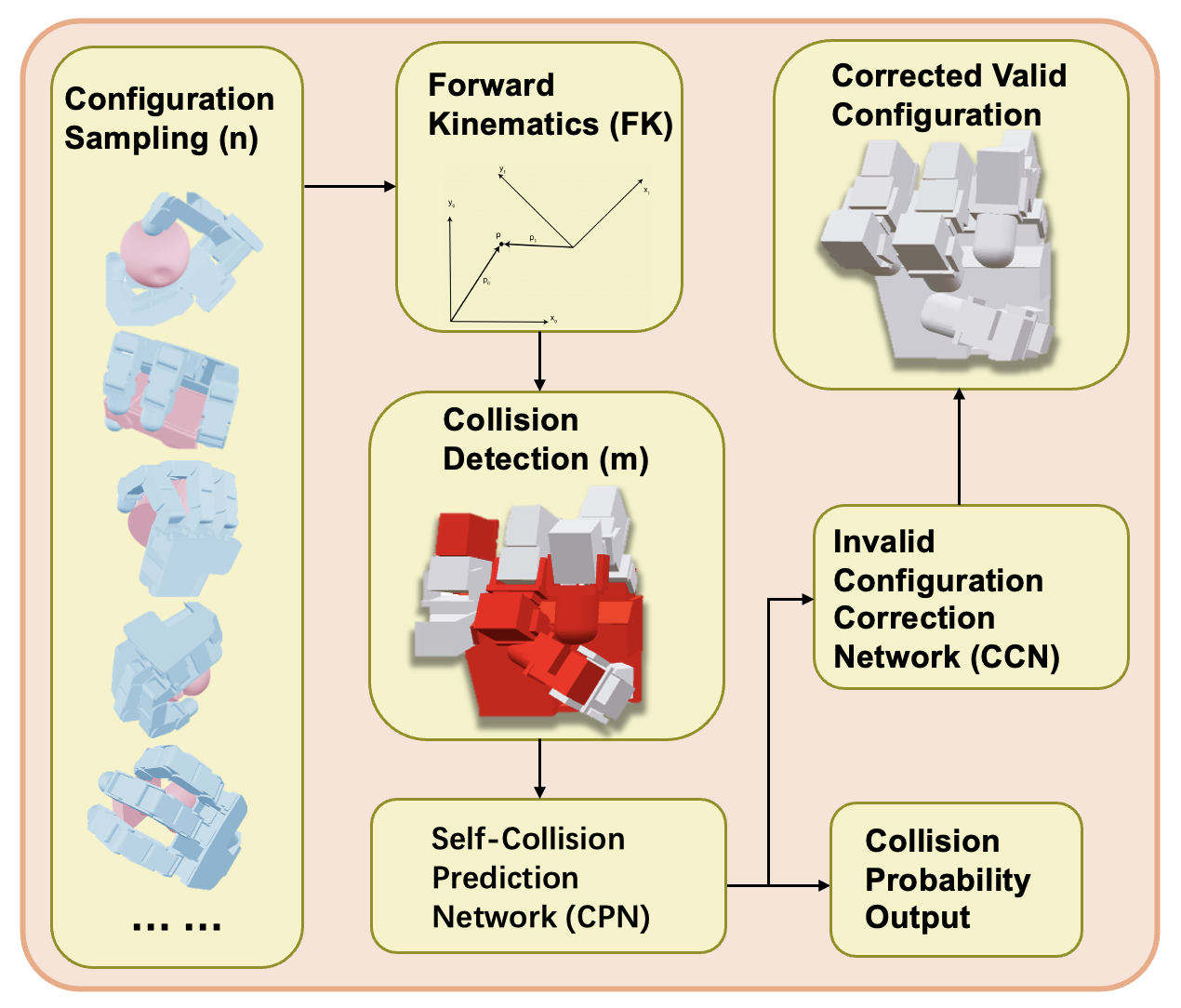}
  \caption{Pipeline of the Non-collision Module}
  \label{fig: flow}
\end{figure}

Fig.~\ref{fig: flow} illustrates our pos2pos implementation pipeline, which consists of several interconnected components for handling robot hand configurations.

\subsubsection{Self-Collision Prediction Network (CPN)}

To facilitate the transformation from invalid to valid configurations, we first develop a \textbf{Self-Collision Prediction Network} (CPN). The primary objective of CPN is to predict the likelihood of self-collision for each link within a given joint configuration.

The training dataset is generated by uniformly sampling $n$ configurations from the robot's action space. For each sampled configuration, the system employs forward kinematics (FK) to compute the robot's pose. A collision detection algorithm (e.g., geometric or physics-based) then checks the pose to derive $m$ collision labels for the links. Each label indicates whether a link is in a collision state.

The CPN takes joint configurations as input and outputs collision probabilities for all joints. We train the network using the binary cross-entropy (BCE) loss function, defined as:
\begin{equation}
    L_{\text{CPN}} = \frac{1}{m} \sum_{i=1}^m \text{BCE}(p_i, t_i),
\end{equation}
where $p_i$ and $t_i$ represent the predicted and true collision probabilities, respectively.

\subsubsection{Invalid Configuration Correction Network (CCN)}

Building on the CPN, we introduce an \textbf{Invalid Configuration Correction Network} (CCN) to map invalid configurations to valid ones. The CCN takes an invalid configuration as input and outputs a corrected configuration that minimizes collision risks while closely resembling the original input.

The CCN training process minimizes a composite loss function comprising two components:
\begin{itemize}
    \item \textbf{Mean Squared Error (MSE) Loss}: Ensures the corrected configuration closely resembles the original configuration.
    \begin{equation}
        L_{\text{MSE}} = \frac{1}{n} \sum_{i=1}^n (\hat{q}_i - q_i)^2,
    \end{equation}
    where $q_i$ and $\hat{q}_i$ denote the original and corrected joint configurations, respectively.
    \item \textbf{Collision Probability Loss}: Leverages the CPN to compute the mean collision probability of the corrected configuration and aims to minimize this value.
    \begin{equation}
        L_{\text{Collision}} = \frac{1}{m} \sum_{i=1}^m p_i(\hat{q}),
    \end{equation}
    where $p_i(\hat{q})$ represents the collision probability of joint $i$ in the corrected configuration $\hat{q}$.
\end{itemize}

We define the total loss function as:
\begin{equation}
    L = \alpha L_{\text{MSE}} + \beta L_{\text{Collision}},
\end{equation}
where $\alpha$ and $\beta$ are hyperparameters balancing the two loss components.

\subsubsection{Explanation and Optimization Strategy}

The loss terms in the proposed framework serve distinct roles:
\begin{itemize}
    \item $L_{\text{MSE}}$ ensures the corrected configuration retains continuity with the original input.
    \item $L_{\text{Collision}}$ minimizes the likelihood of self-collision in the corrected configuration.
    \item The hyperparameters $\alpha$ and $\beta$ significantly influence the training outcomes, and we optimize their values through grid search.
\end{itemize}

The CCN employs a fully connected multi-layer perceptron (MLP) architecture. The input is the invalid joint configuration $q$, and the output is the corrected configuration $\hat{q}$. We train the model using the Adam optimizer with a learning rate $\eta$ and monitor convergence via the collision rate on a validation dataset.

\subsubsection{Summary}

By integrating the CPN and CCN, we efficiently transform invalid self-collision configurations into valid ones. This approach ensures the validity and continuity of robotic configurations, laying a robust foundation for subsequent task execution.

\subsection{Bill of Materials (BOM)}

Our teleoperation system supports a modular architecture with flexible input configurations. \textbf{Not all devices shown in Fig. \ref{fig: framework} are required simultaneously.} Instead, the system requires \textbf{one device from each of the two input groups} on the left side of the diagram:

\begin{itemize}
    \item \textbf{Wrist pose acquisition:} RGB(-D) camera, IMU mocap suit, or similar.
    \item \textbf{Hand gesture acquisition:} Mocap gloves, AR tracking, or EMG-based sensing.
\end{itemize}

This design allows users to build their system using available hardware, optimizing for cost, performance, or ease of use. For example, while a VR headset can serve as a unified sensor in both categories, it is not necessary. Our system can be operated using a standard external monitor, as done in our experiments.

\subsubsection*{Experimental Setup Used in This Paper}

In our experiments, we selected a configuration based on performance, generalizability, and affordability:

\begin{itemize}
    \item \textbf{Mocap Gloves:} \$500  
    \begin{itemize}
        \item \href{https://www.kickstarter.com/projects/udexreal/udcap-silk-like-vr-gloves-for-steamvr}{Kickstarter Product Page}
    \end{itemize}

    \item \textbf{IMU-Based Motion Capture Suit:} \$200  
    \begin{itemize}
        \item \href{https://store.rebocap.site/products/rebocap-15-point-full-body-inertial-motion-capture}{Rebocap Product Page}
    \end{itemize}

    \item \textbf{RGB-D Camera (Intel RealSense D435):} \$300  
    \begin{itemize}
        \item \href{https://www.amazon.com/Intel-Realsense-D435-Webcam-FPS/dp/B07BLS5477}{Amazon Product Page}
    \end{itemize}

    \item \textbf{Total Cost:} Approximately \textbf{\$1000}
\end{itemize}

\textit{Note:} Since our method does not depend on depth data, the RGB-D camera can be replaced by a lower-cost RGB camera, further reducing the overall system cost.

\subsubsection*{Remarks on Reproducibility and Flexibility}

The modular nature of the TelePreview architecture allows for hardware substitution based on availability or task requirements. We provide this BOM to facilitate future replication efforts and to emphasize that \textbf{our system's affordability claim refers to the teleoperation input interface only}, not the robot arm or end-effector hardware.

\subsection{Reproduction Experience of Baseline Methods}

We tested several typical vision-based teleoperation methods under our experimental setup: OpenTeach, OpenTelevision, and AnyTeleop. For methods that did not support LeapHand in the open-source code, we implemented the corresponding parts ourselves to apply these methods.

\textbf{OpenTeach:} Following the guidelines in the open-source repository for adding new hardware, we implemented the retargeting module for LeapHand. However, we encountered two significant issues. First, gesture recognition based on visual input is highly sensitive to occlusions. When the back of the hand is fully occluded or the side of the hand is partially blocked, the positioning of the fingers can deviate substantially—particularly with the Meta Quest3, which has less accurate hand tracking compared to the Apple Vision Pro. Second, the visual approach depends heavily on precise hand calibration, requiring careful adjustment of parameters to achieve acceptable retargeting results. Together, these two factors led to instability in the accuracy of teleoperation in some instances, ultimately impacting the success rate of task completion.

\textbf{OpenTelevision:} The open-source code only provides the necessary components for the 6-DoF Inspire Hand, with no guidance on how to integrate new dexterous hands. Additionally, many hand-specific hyperparameters lack proper documentation. After attempting to integrate LeapHand using the dex-retargetting code and redoing the joint mapping, we were able to align the retargeting results with the hand's movements. However, due to LeapHand's significantly higher degrees of freedom, we encountered severe self-collision issues. During deployment, motor collisions occurred, rendering effective teleoperation impossible. Such self-collision problems did not arise with the original Inspire Hand, as its 6-DoF design inherently prevents the possibility of self-collisions.

\textbf{AnyTeleop:} The retargeting code used here is also based on dex-retargetting. For LeapHand, a high-DOF dexterous hand with a mechanical design that has a higher likelihood of self-collision, relying solely on fingertip-based inverse kinematics (IK) resulted in frequent self-collisions. This posed significant issues during teleoperation.

\subsection{User Study in Subjective and Practice Time Evaluation}

To complement the quantitative performance evaluation, we conducted a user study measuring both subjective workload and practice time across different input conditions. Our goal was to assess how the preview system impacts perceived difficulty and actual learning effort.

\subsubsection{Subjective Workload Ratings}

We asked participants to rate their experience using a 5-point Likert scale (higher values indicate greater intensity) across five dimensions: \textit{Mental Demand}, \textit{Physical Demand}, \textit{Total Demand}, \textit{Frustration Level}, and \textit{Perceived Performance}. Each participant completed all tasks under both \textbf{w/o Preview} and \textbf{w/ Preview} conditions.

\begin{figure}[ht]
    \centering
    \includegraphics[width=0.9\linewidth]{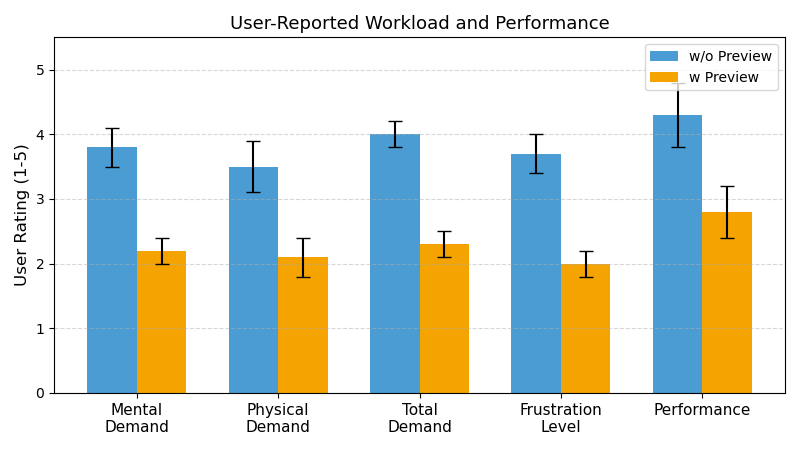}
    \caption{User-reported workload and performance across five dimensions using in \cite{li2025trainrobotsimpactdemonstration}. Ratings were collected using a 5-point Likert scale.}
    \label{fig: workload}
    \vspace{-15pt}
\end{figure}

As shown in Figure~\ref{fig: workload}, the use of the preview system significantly reduced user-reported workload across all categories. In particular, users reported substantially lower mental and physical demand and improved overall performance perception under the \textbf{w/ Preview} condition. This indicates that the preview feature not only improves task success but also reduces cognitive and physical strain.

\subsubsection{Practice Time to Confidence}

We also recorded the time participants spent practicing each control modality until they reported being comfortable with beginning formal task trials. Three input conditions were compared: our proposed method \textbf{w/ Preview}, the same setup \textbf{w/o Preview}, and a baseline \textbf{vision-based method}\cite{qin2023anyteleop} commonly used in prior work.

\begin{figure}[ht]
    \centering
    \includegraphics[width=0.8\linewidth]{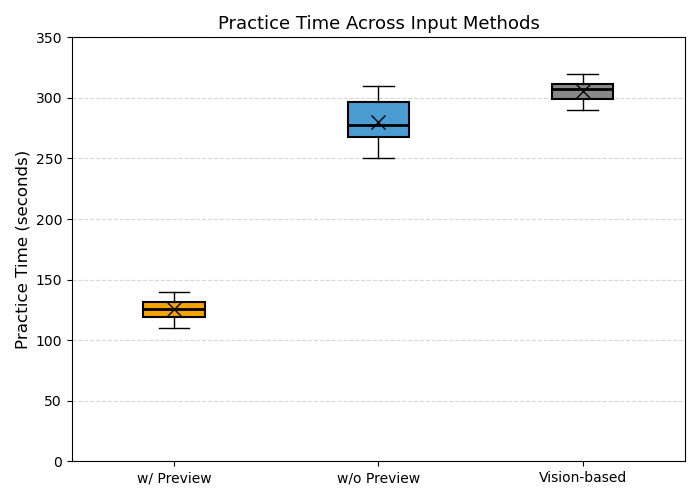}
    \caption{Practice time. New users reached task readiness faster with our system (w/ Preview) than with other input methods.}
    \label{fig: practice}
\end{figure}

As shown in Figure~\ref{fig: practice}, participants required the least time to reach confidence using our full TelePreview setup. The vision-based baseline took the longest and also showed the largest variability across users. These results support the claim that previewed, body-mapped teleoperation enables faster and more intuitive learning.

\subsection{End-Effector Integration Details}
\label{appendix:end_effector_results}

\begin{figure}[htbp]
\centering
\begin{subfigure}{0.4\linewidth}
   \centering
   \includegraphics[width=\linewidth]{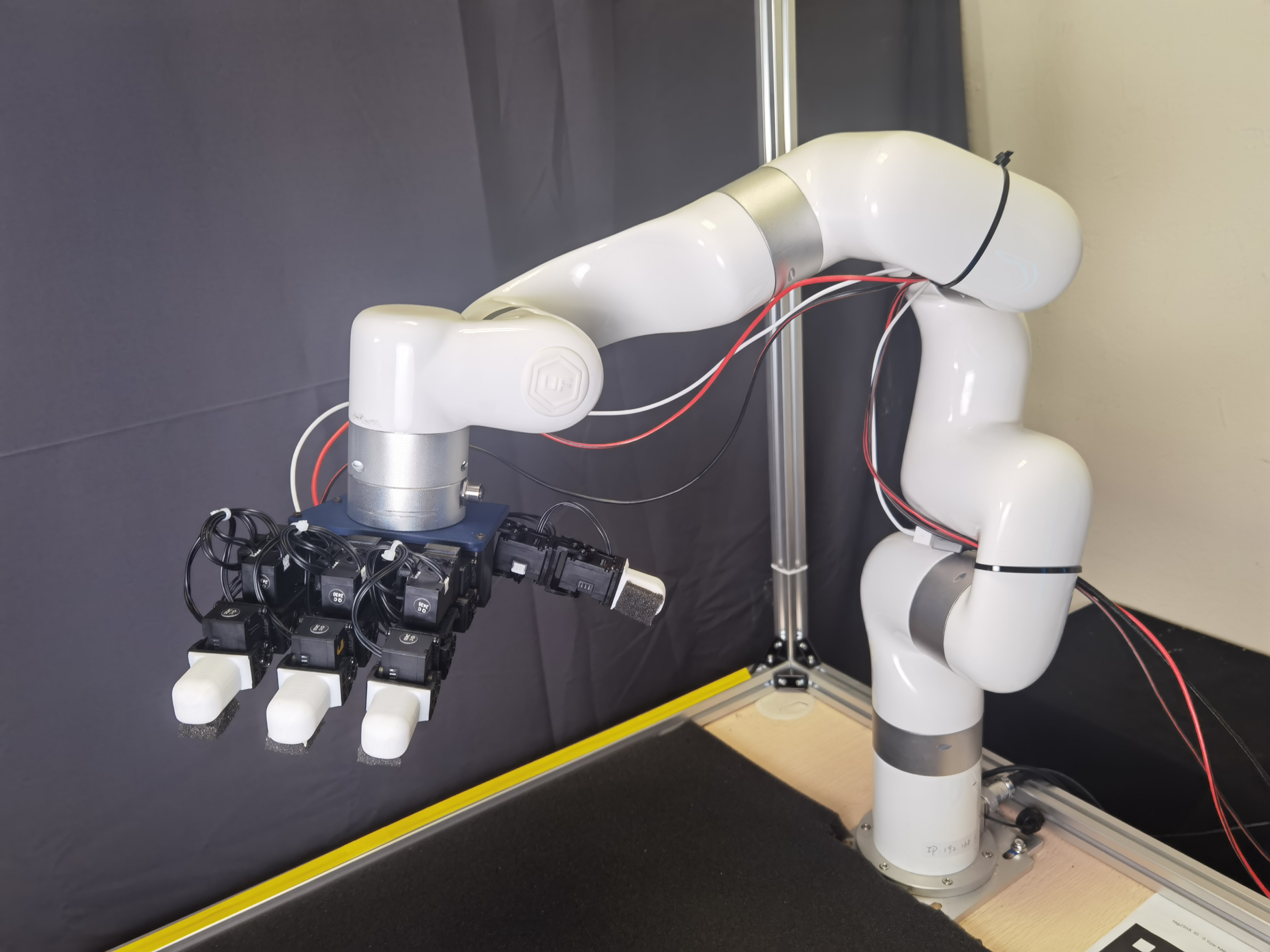}
   \caption{LEAP Hand}
   \label{fig: hand}
\end{subfigure}
\begin{subfigure}{0.4\linewidth}
   \centering
   \includegraphics[width=\linewidth]{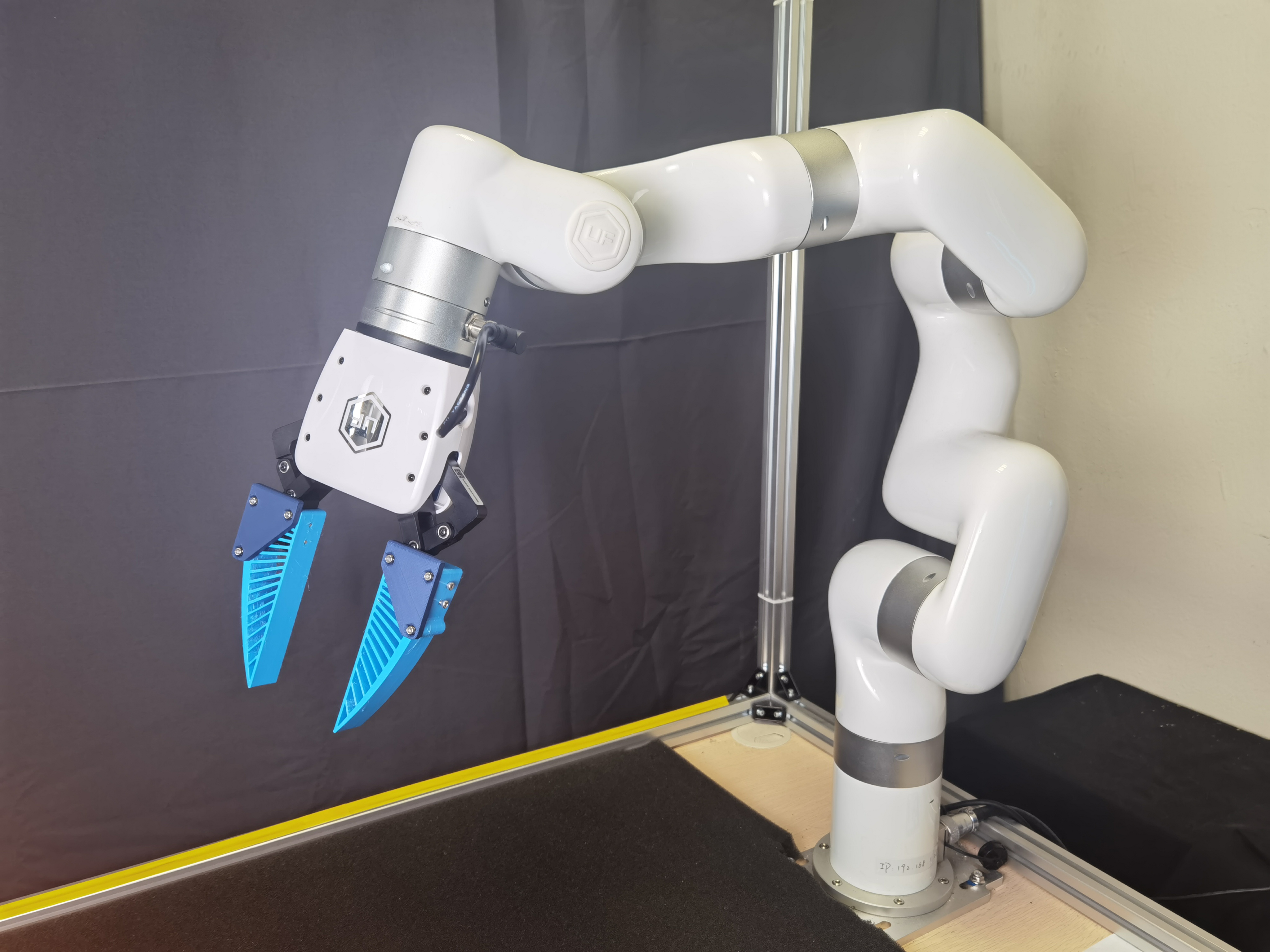}
   \caption{Gripper}
   \label{fig: gripper}
\end{subfigure}
\caption{Deployment on Different Robots.}
\label{fig: deploy}
\end{figure}

To demonstrate the generalizability of our system, we deployed TelePreview on a Ufactory xArm robot equipped with three different end-effectors(See in Fig. \ref{fig: deploy}):

\begin{itemize}
    \item \textbf{LeapHand (multi-DoF anthropomorphic hand)}: Controlled via direct mapping of 16 captured joint angles from the mocap glove through our SMPL-X based retargeting pipeline. The preview and execution follow the full configuration space of the robot hand, enabling rich dexterous behaviors.
    
    \item \textbf{Parallel-jaw Gripper (single-DoF)}: We selected a representative finger joint angle from the mocap glove and used its value to control the gripper's open/close motion. This allowed the system to retain the preview feature without the need for full-hand mapping.
    
    \item \textbf{Vacuum Gripper (binary actuator)}: We applied a threshold to the same glove joint signal to produce a binary on/off activation, simulating ``grasp'' or ``release'' behavior. This minimal control model was still compatible with the preview system.

\end{itemize}

In all three configurations, only minor parameter adjustments were required (e.g., kinematic model, end-effector transform), and the core TelePreview architecture remained unchanged. These results support our claim that the system is hardware-agnostic and can be adapted to different robot platforms with minimal integration effort.

\begin{table}[ht]
\centering
\resizebox{\columnwidth}{!}{%
\begin{tabular}{lccc}
\toprule
\textbf{End-Effector} & \textbf{w/o Preview} & \textbf{w/ Preview} & \textbf{Improvement} \\
\midrule
LeapHand (16-DoF)             & $23.6 \pm 4.7$  & $13.6 \pm 3.2$   & $-10.0$ \\
Parallel-jaw Gripper (1-DoF)  & $16.8 \pm 1.7$  & $14.2 \pm 1.4$ & $-2.6$  \\
Vacuum Gripper (binary)       & $15.2 \pm 1.9$  & $13.7 \pm 1.3$ & $-1.5$  \\
\bottomrule
\end{tabular}%
}
\caption{Average execution time (in seconds) across different end-effectors with and without the preview feature in the Pick \& Place task. TelePreview shows the most benefit on the high-DoF LeapHand.}
\label{tab:end_effector_results}
\end{table}

As shown in Table~\ref{tab:end_effector_results}, the preview system yields the greatest execution time improvement with the LeapHand, supporting our claim that TelePreview is most beneficial for high-DoF, complex manipulation tasks.

\subsection{User Evaluation Questionnaire}
\label{appendix:user-questionnaire}

After completing the teleoperation tasks under each input condition, participants were asked to rate their experience across five categories using a 5-point Likert scale. Higher scores indicate greater intensity unless otherwise specified.

\begin{table}[ht]
\centering
\small
\begin{tabular}{|p{0.36\linewidth}|p{0.55\linewidth}|}
\hline
\textbf{Question} & \textbf{Rating Scale (1–5)} \\
\hline
How mentally demanding was the task? & 1 = Very Low, 5 = Very High \\
\hline
How physically demanding was the task? & 1 = Very Low, 5 = Very High \\
\hline
How frustrated did you feel while using the system? & 1 = Not at all, 5 = Extremely \\
\hline
How much overall effort did the task require? & 1 = Very Little, 5 = Very Much \\
\hline
How successful do you feel you were in completing the tasks? & 1 = Very Unsuccessful, 5 = Very Successful \\
\hline
\end{tabular}
\caption{Likert-scale questionnaire completed after each input condition.}
\label{tab:user-questionnaire}
\end{table}

\vspace{-0.5cm}
\noindent
\textit{
\textbf{Mentally demanding}: The degree of cognitive effort required (e.g., concentration, decision-making). \\
\textbf{Physically demanding}: The amount of physical exertion needed (e.g., hand or body movement, fatigue). \\
\textbf{Frustration}: The extent of annoyance, stress, or irritation experienced. \\
\textbf{Overall effort}: The perceived total effort needed to perform the task, combining physical and mental demands.
}

Participants answered this questionnaire once for each input modality they used (e.g., \textit{w/ Preview}, \textit{w/o Preview}).

\subsection{Visualization of our tasks}
\begin{figure*}[htbp]
    \centering
    \begin{subfigure}[b]{0.18\textwidth}
        \centering
        \includegraphics[width=\textwidth]{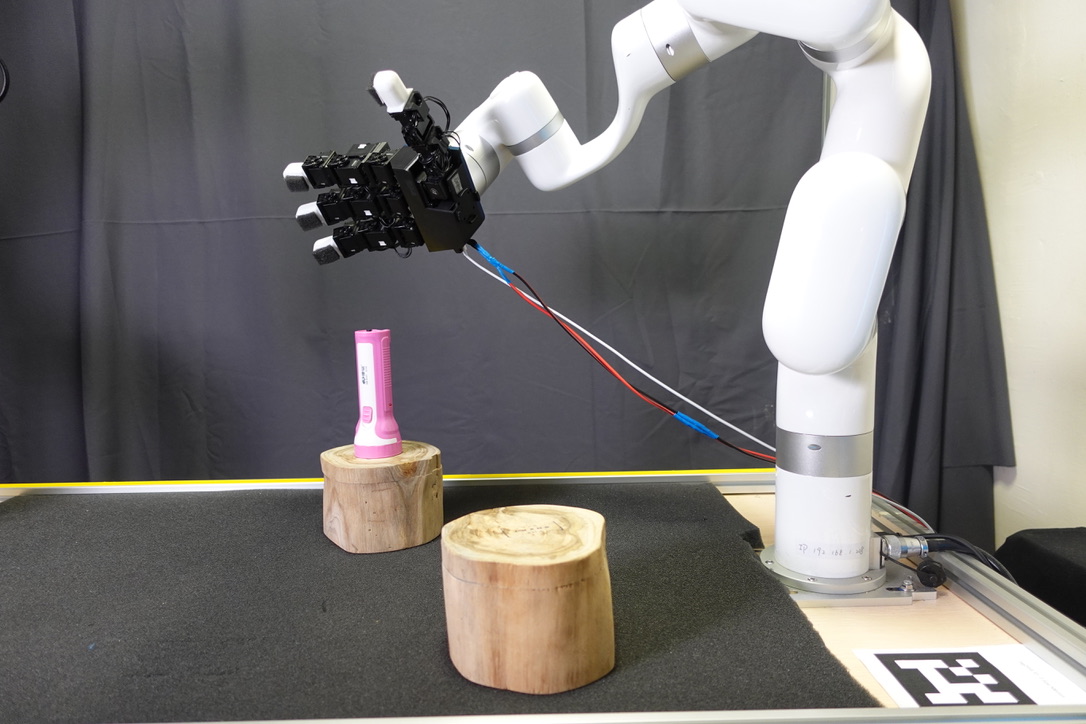}
        \caption{}
    \end{subfigure}
    \begin{subfigure}[b]{0.18\textwidth}
        \centering
        \includegraphics[width=\textwidth]{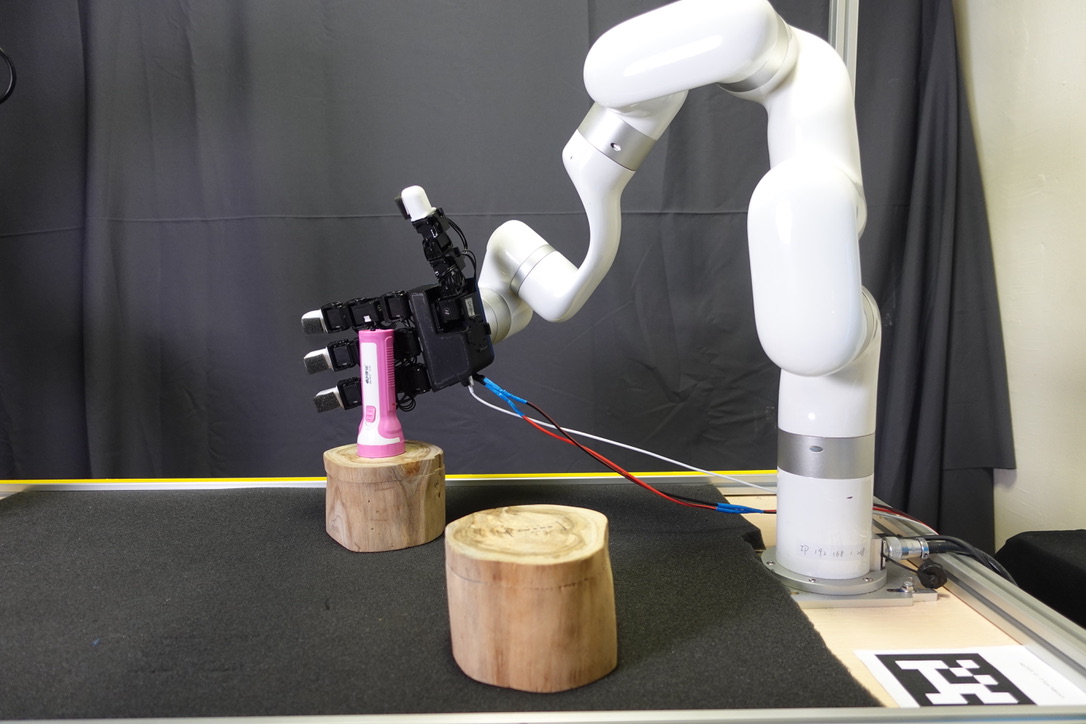}
        \caption{}
    \end{subfigure}
    \begin{subfigure}[b]{0.18\textwidth}
        \centering
        \includegraphics[width=\textwidth]{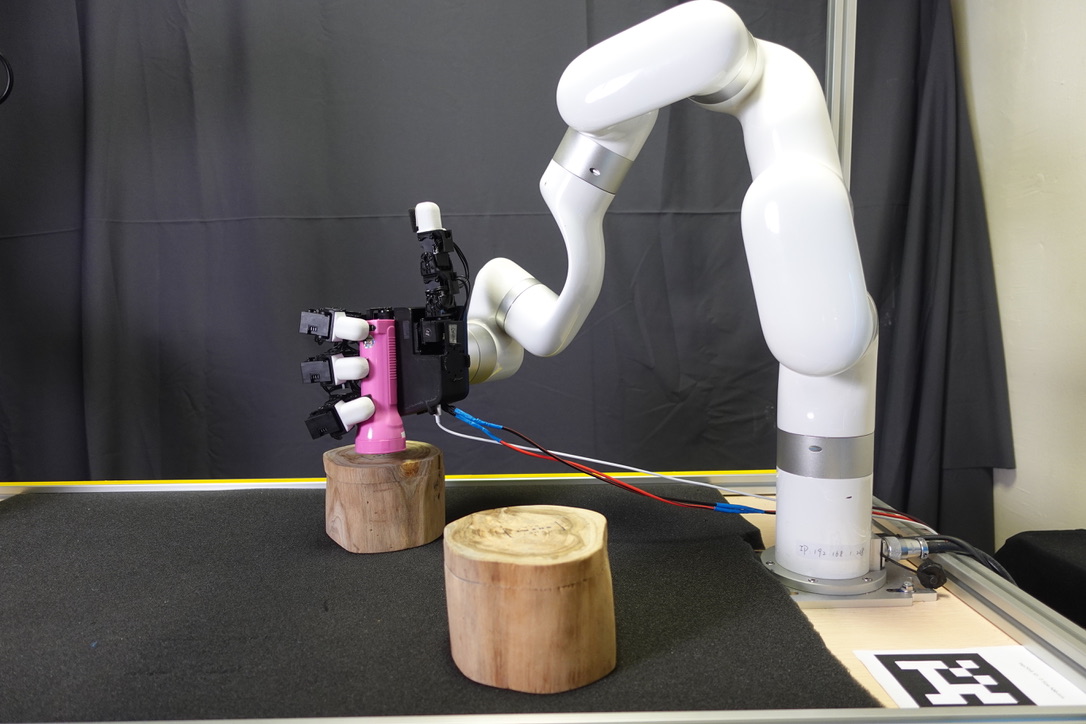}
        \caption{}
    \end{subfigure}
    \begin{subfigure}[b]{0.18\textwidth}
        \centering
        \includegraphics[width=\textwidth]{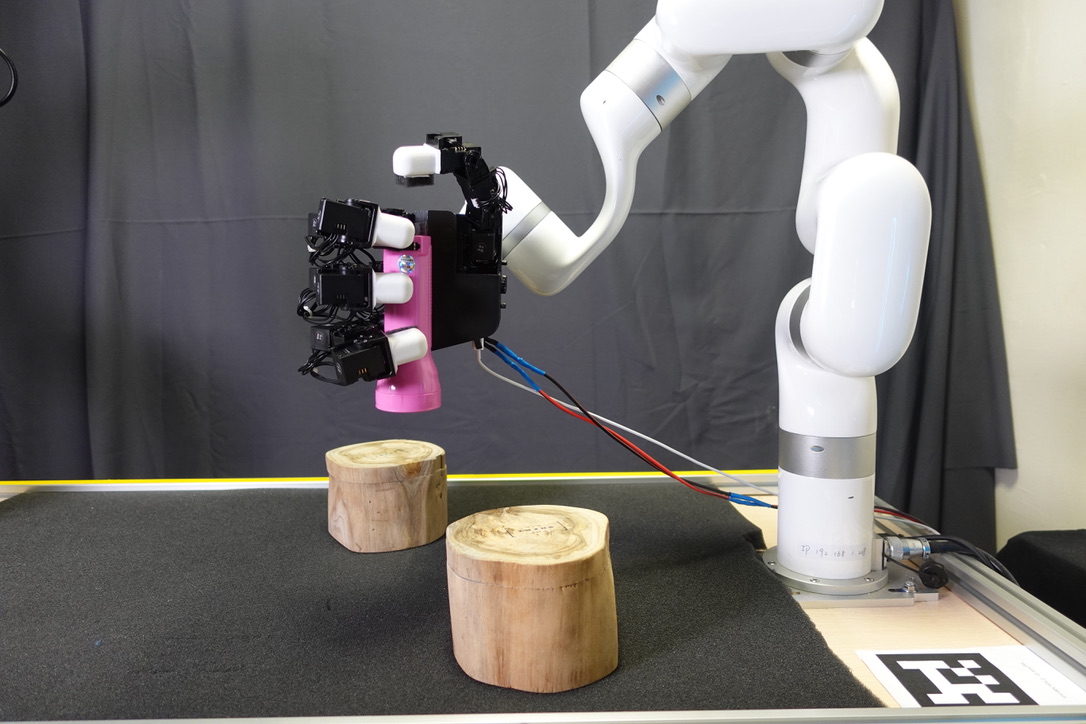}
        \caption{}
    \end{subfigure}
    \begin{subfigure}[b]{0.18\textwidth}
        \centering
        \includegraphics[width=\textwidth]{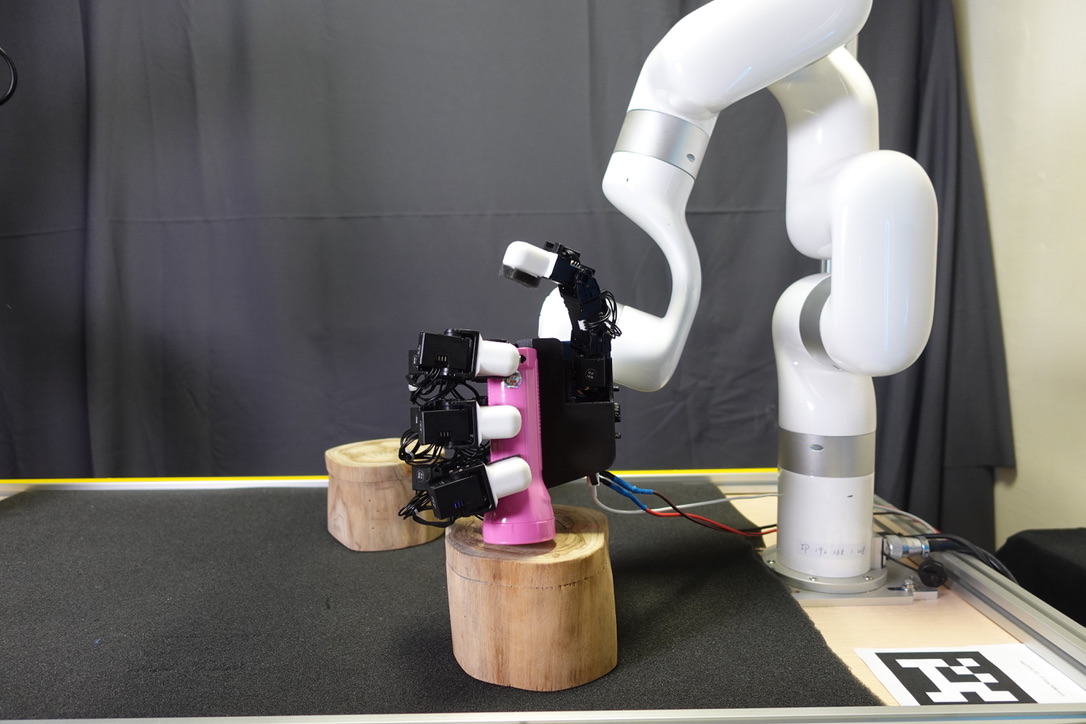}
        \caption{}
    \end{subfigure}
    \caption{Pick\&Place Visualization}
    \label{fig: PickPlace}
\end{figure*}

\begin{figure*}[htbp]
    \centering
    \begin{subfigure}[b]{0.18\textwidth}
        \centering
        \includegraphics[width=\textwidth]{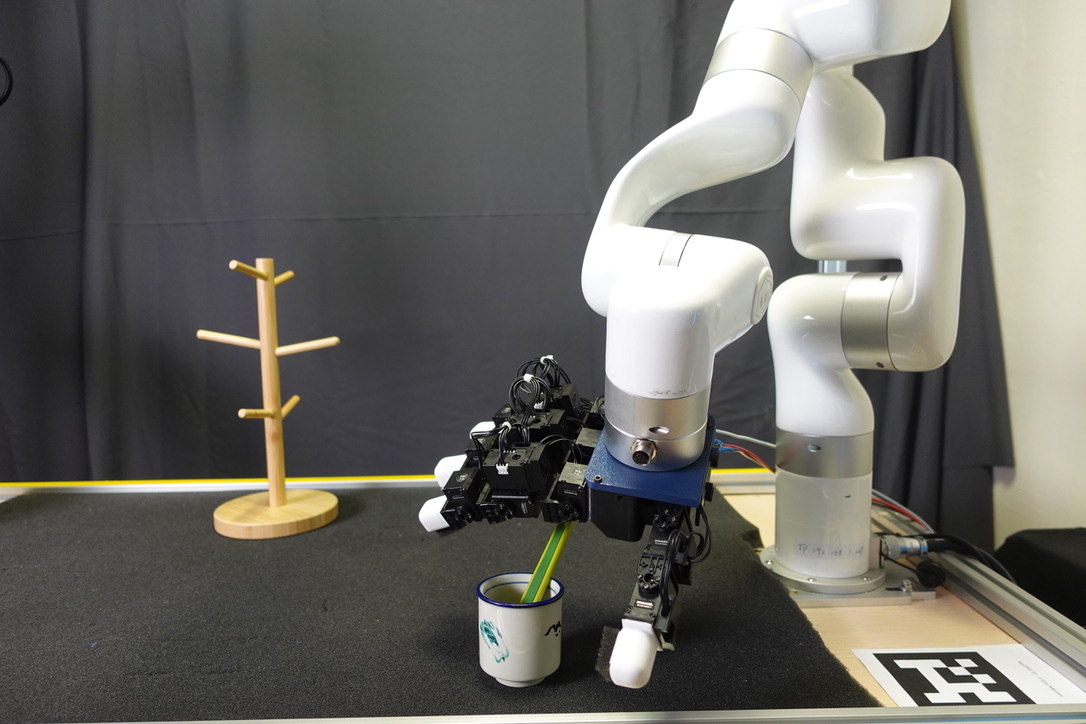}
        \caption{}
    \end{subfigure}
    \begin{subfigure}[b]{0.18\textwidth}
        \centering
        \includegraphics[width=\textwidth]{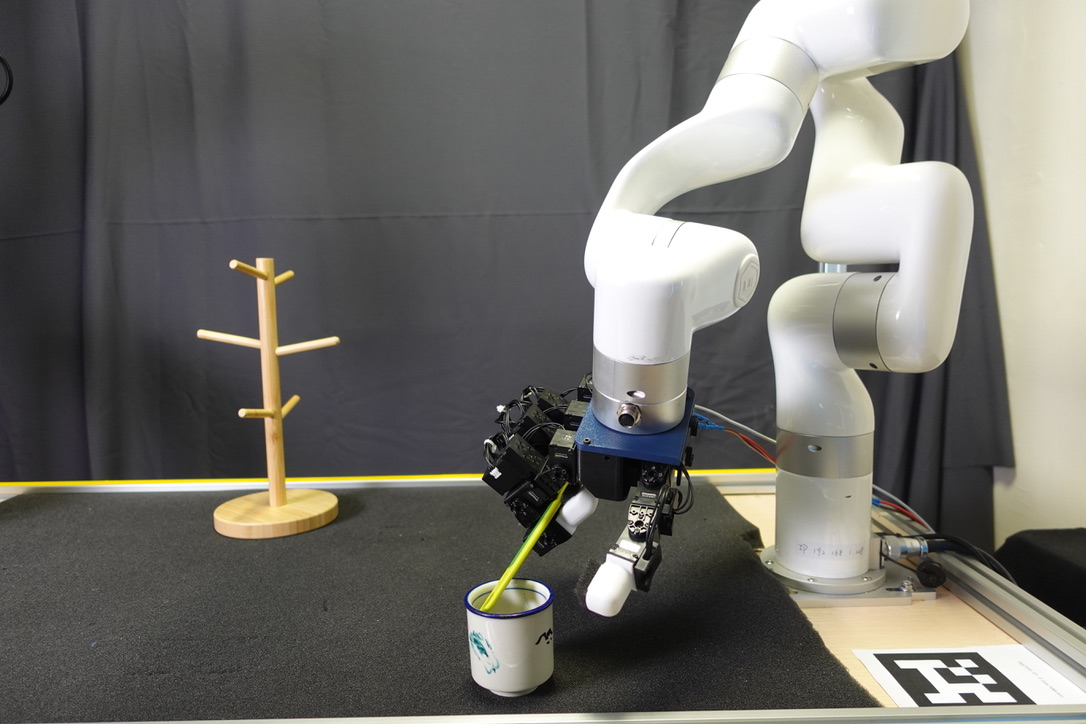}
        \caption{}
    \end{subfigure}
    \begin{subfigure}[b]{0.18\textwidth}
        \centering
        \includegraphics[width=\textwidth]{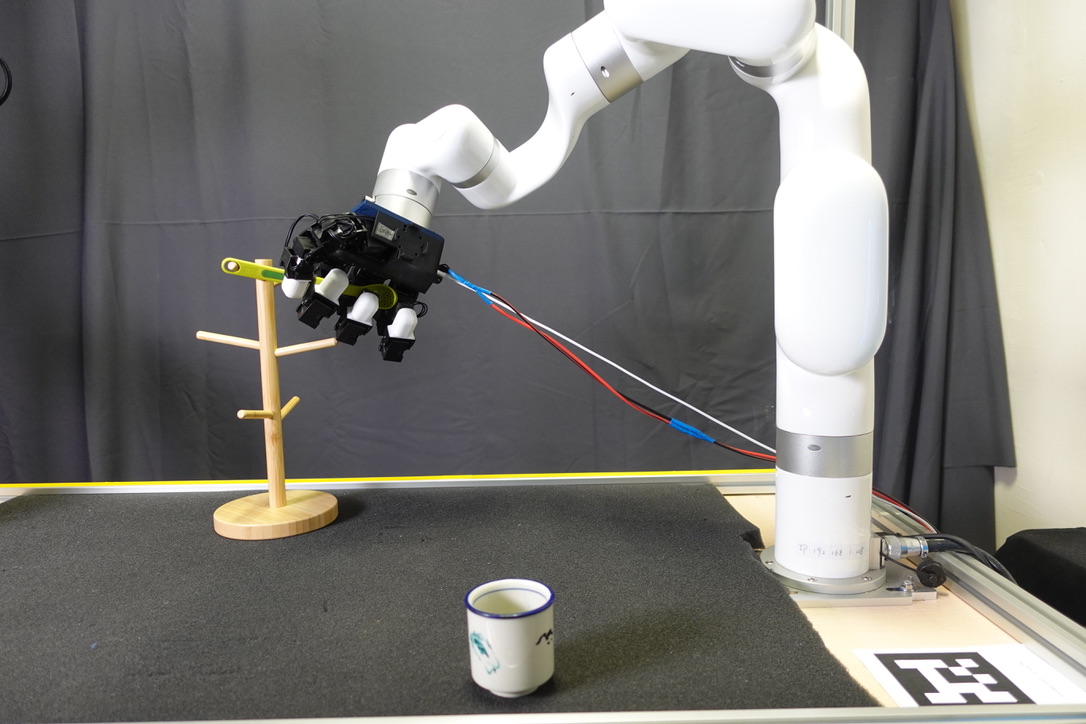}
        \caption{}
    \end{subfigure}
    \begin{subfigure}[b]{0.18\textwidth}
        \centering
        \includegraphics[width=\textwidth]{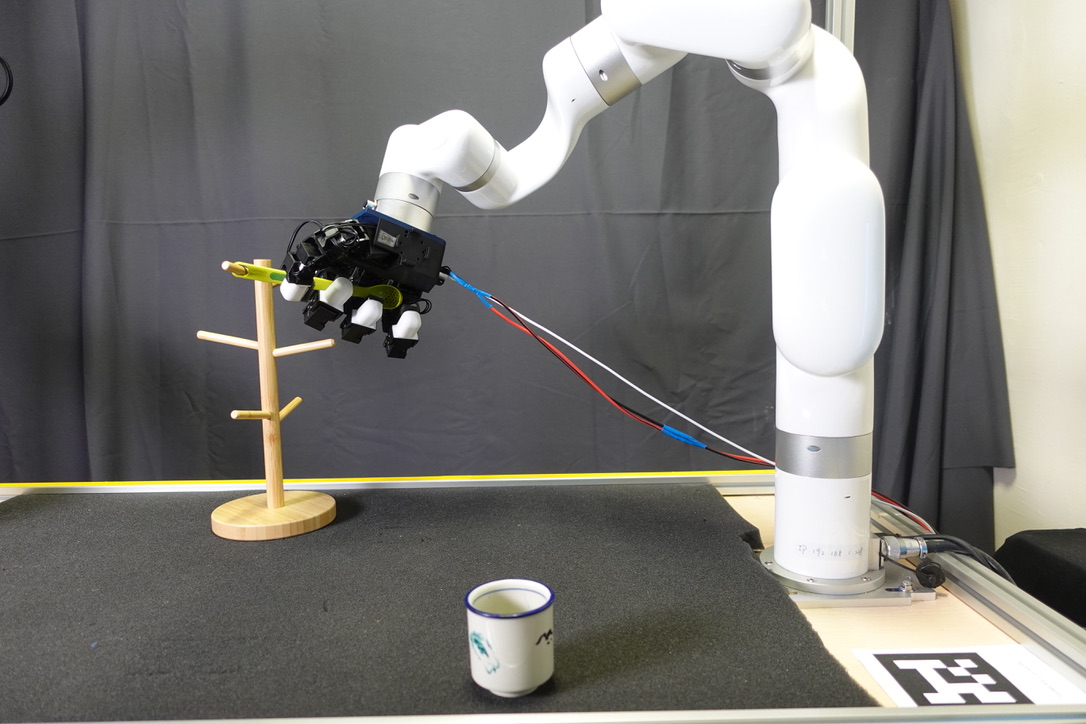}
        \caption{}
    \end{subfigure}
    \begin{subfigure}[b]{0.18\textwidth}
        \centering
        \includegraphics[width=\textwidth]{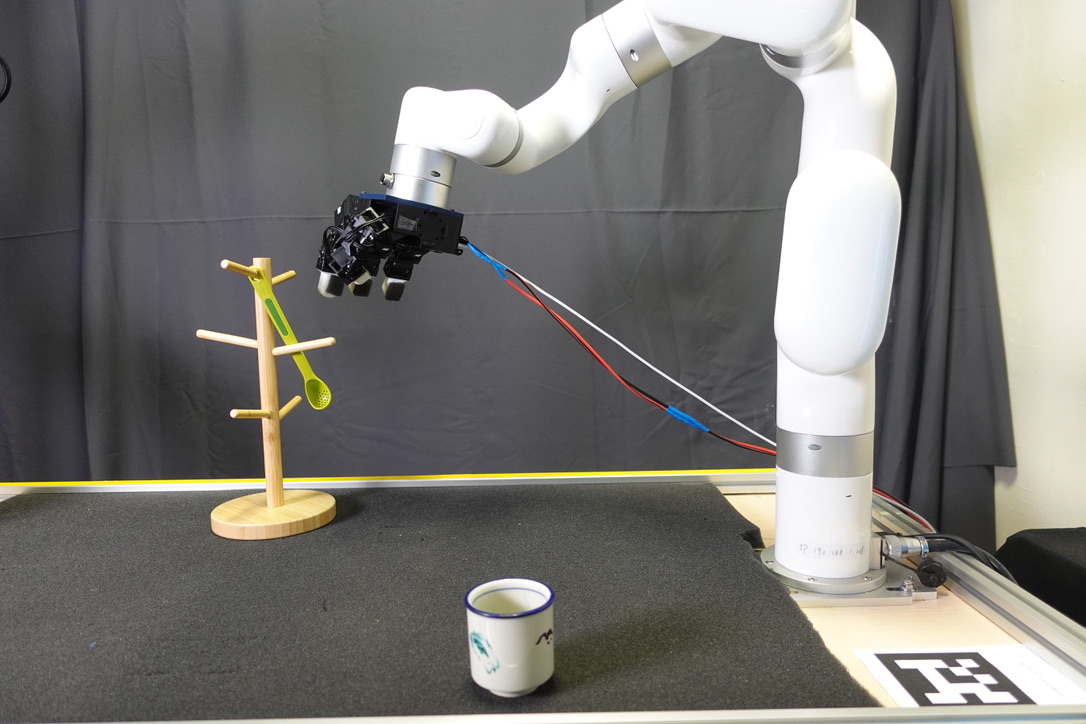}
        \caption{}
    \end{subfigure}
    \caption{Hang Visualization}
    \label{fig:hang}
\end{figure*}

\begin{figure*}[htbp]
    \centering
    \begin{subfigure}[b]{0.18\textwidth}
        \centering
        \includegraphics[width=\textwidth]{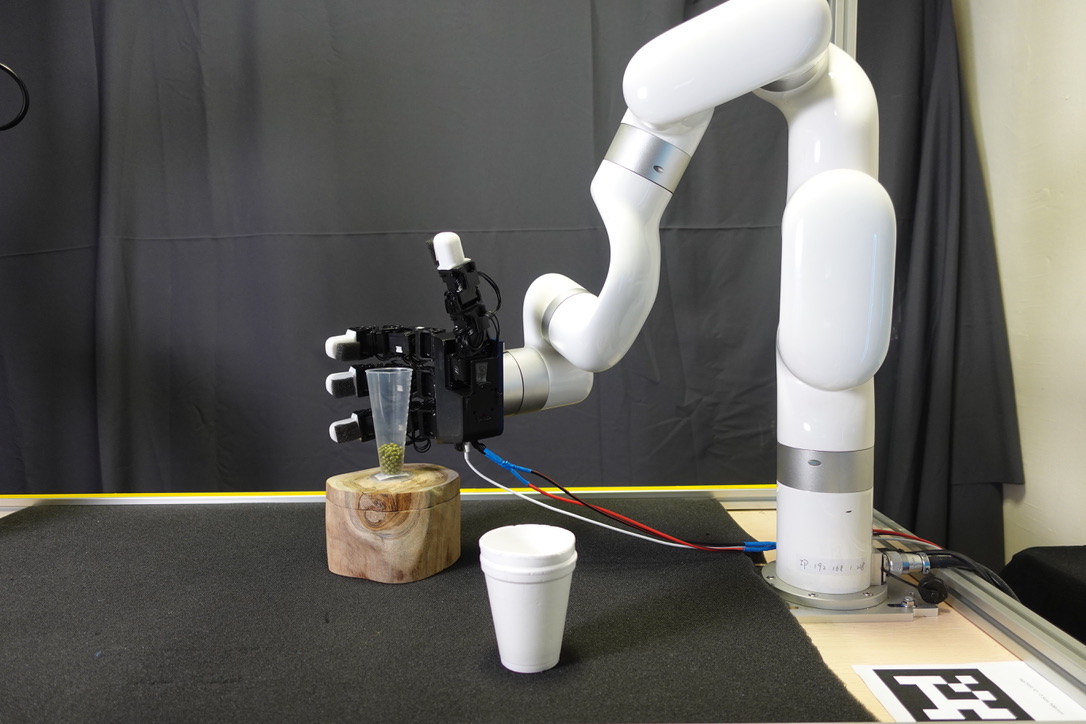}
        \caption{}
    \end{subfigure}
    \begin{subfigure}[b]{0.18\textwidth}
        \centering
        \includegraphics[width=\textwidth]{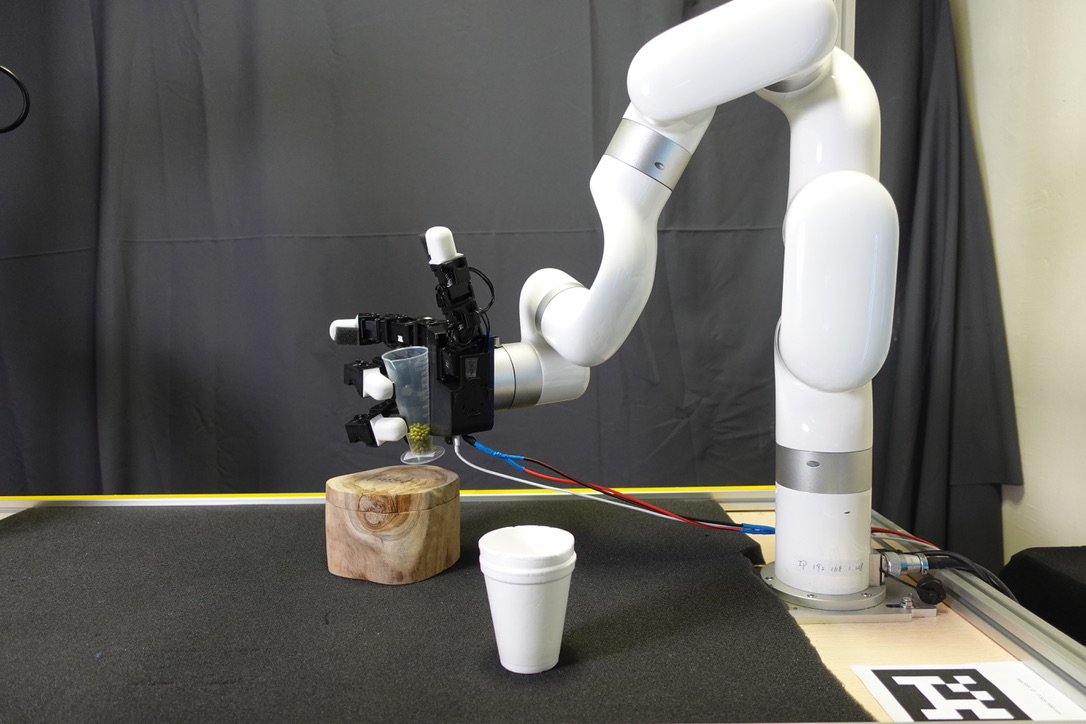}
        \caption{}
    \end{subfigure}
    \begin{subfigure}[b]{0.18\textwidth}
        \centering
        \includegraphics[width=\textwidth]{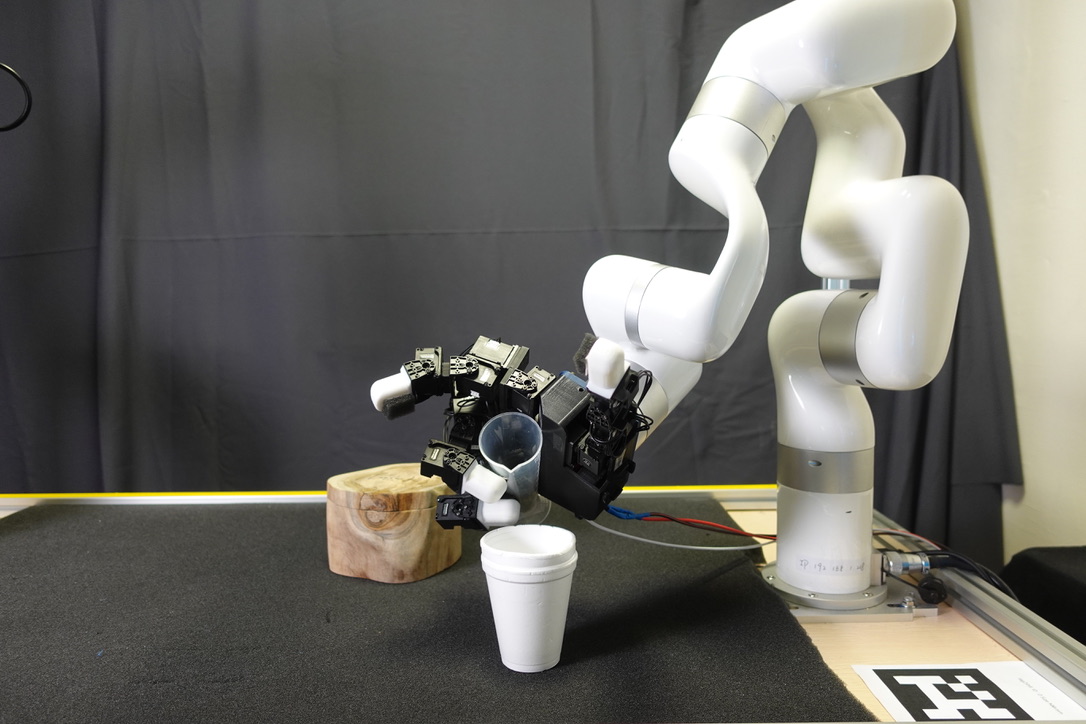}
        \caption{}
    \end{subfigure}
    \begin{subfigure}[b]{0.18\textwidth}
        \centering
        \includegraphics[width=\textwidth]{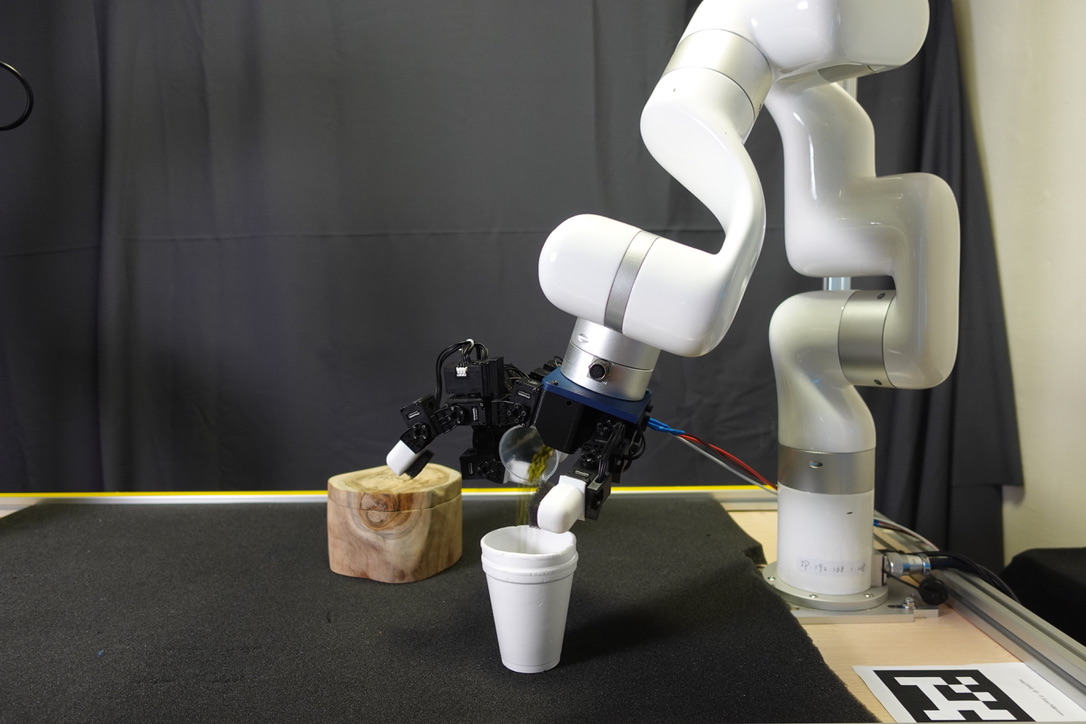}
        \caption{}
    \end{subfigure}
    \begin{subfigure}[b]{0.18\textwidth}
        \centering
        \includegraphics[width=\textwidth]{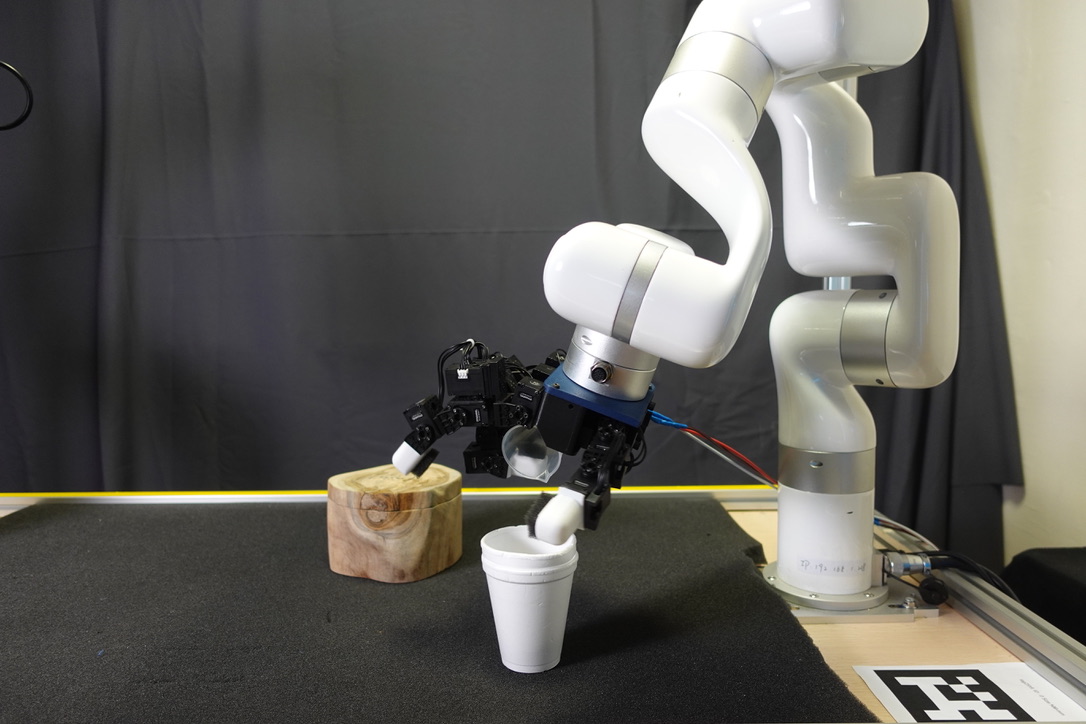}
        \caption{}
    \end{subfigure}
    \caption{Pour Visualization}
    \label{fig:pour}
\end{figure*}

\begin{figure*}[htbp]
    \centering
    \begin{subfigure}[b]{0.18\textwidth}
        \centering
        \includegraphics[width=\textwidth]{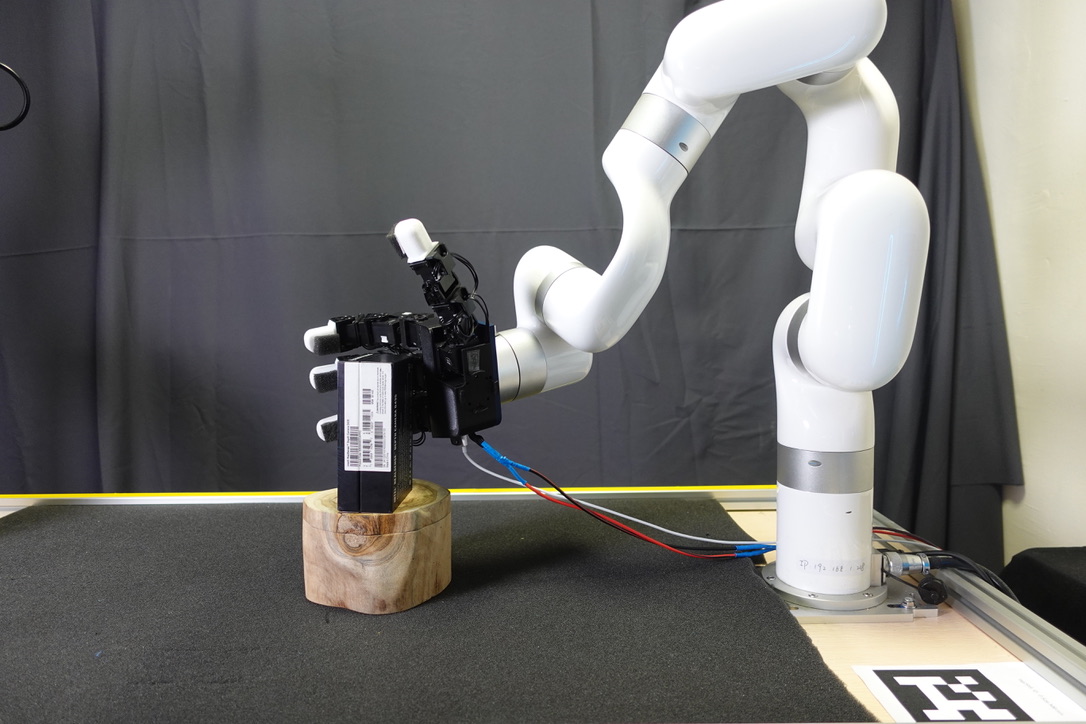}
        \caption{}
    \end{subfigure}
    \begin{subfigure}[b]{0.18\textwidth}
        \centering
        \includegraphics[width=\textwidth]{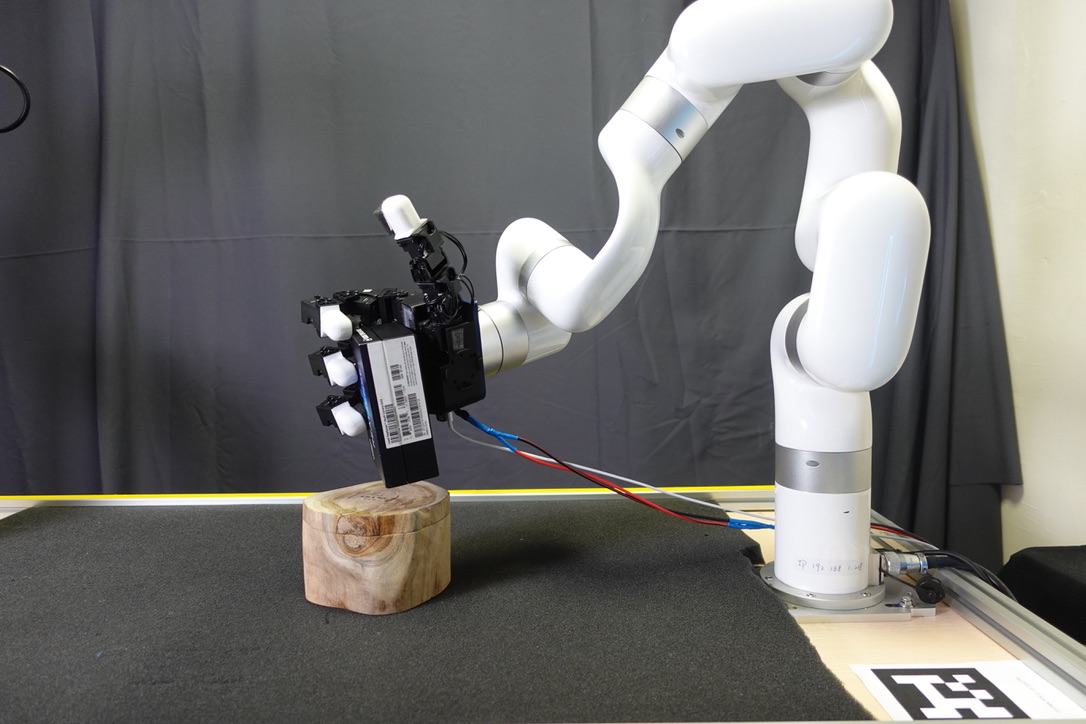}
        \caption{}
    \end{subfigure}
    \begin{subfigure}[b]{0.18\textwidth}
        \centering
        \includegraphics[width=\textwidth]{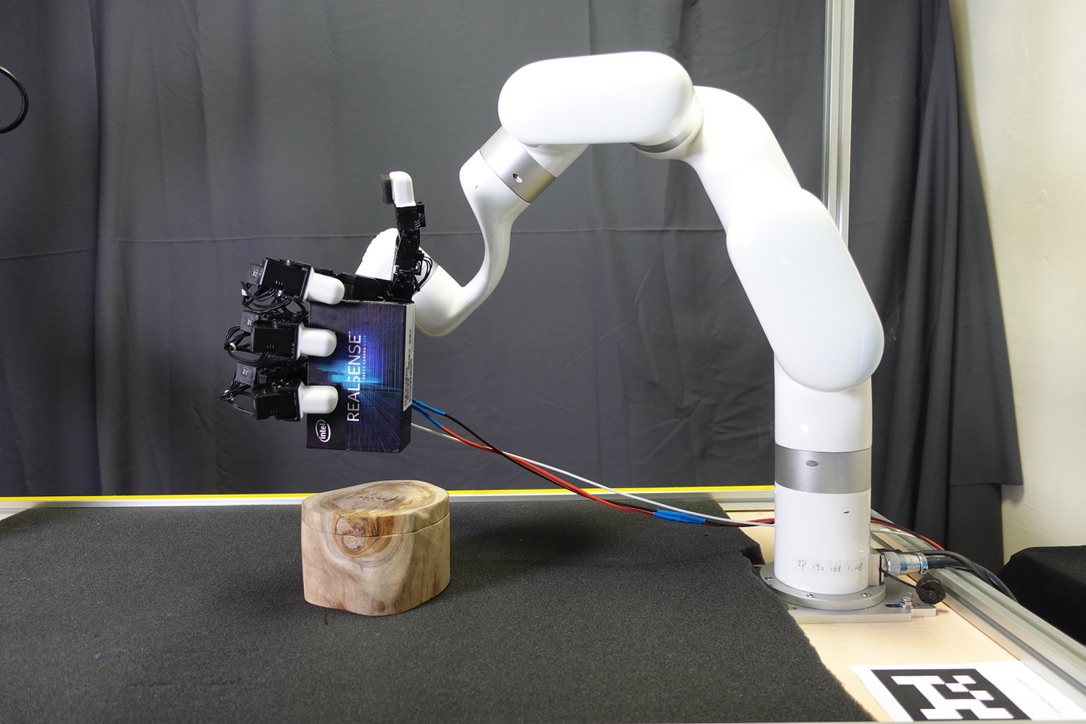}
        \caption{}
    \end{subfigure}
    \begin{subfigure}[b]{0.18\textwidth}
        \centering
        \includegraphics[width=\textwidth]{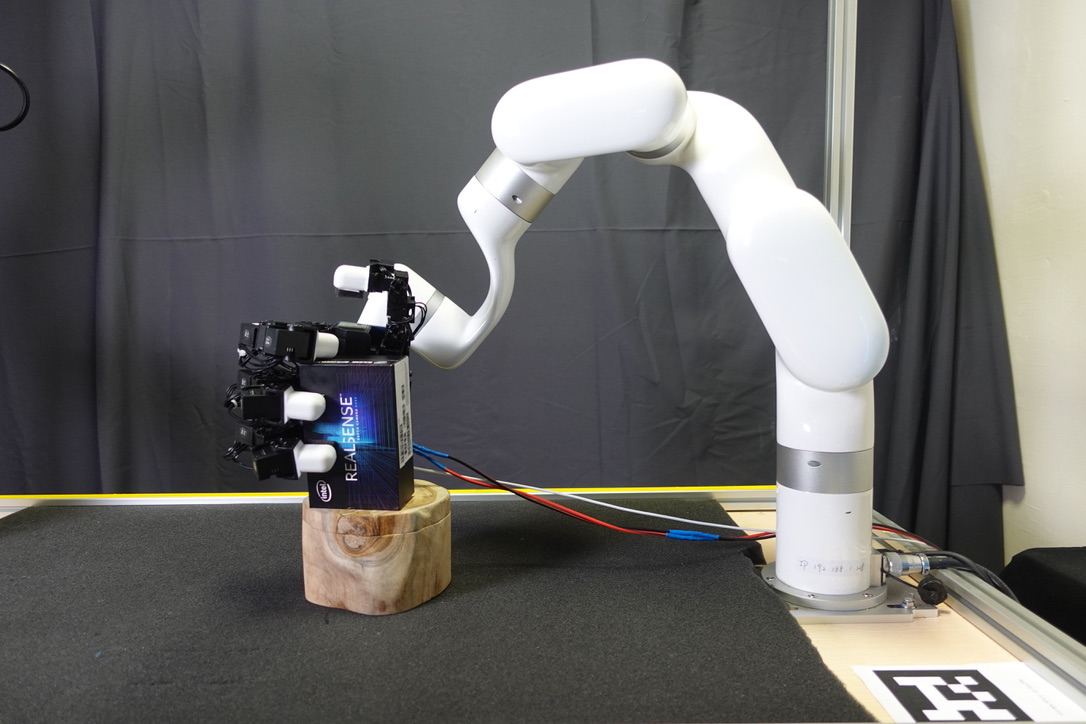}
        \caption{}
    \end{subfigure}
    \begin{subfigure}[b]{0.18\textwidth}
        \centering
        \includegraphics[width=\textwidth]{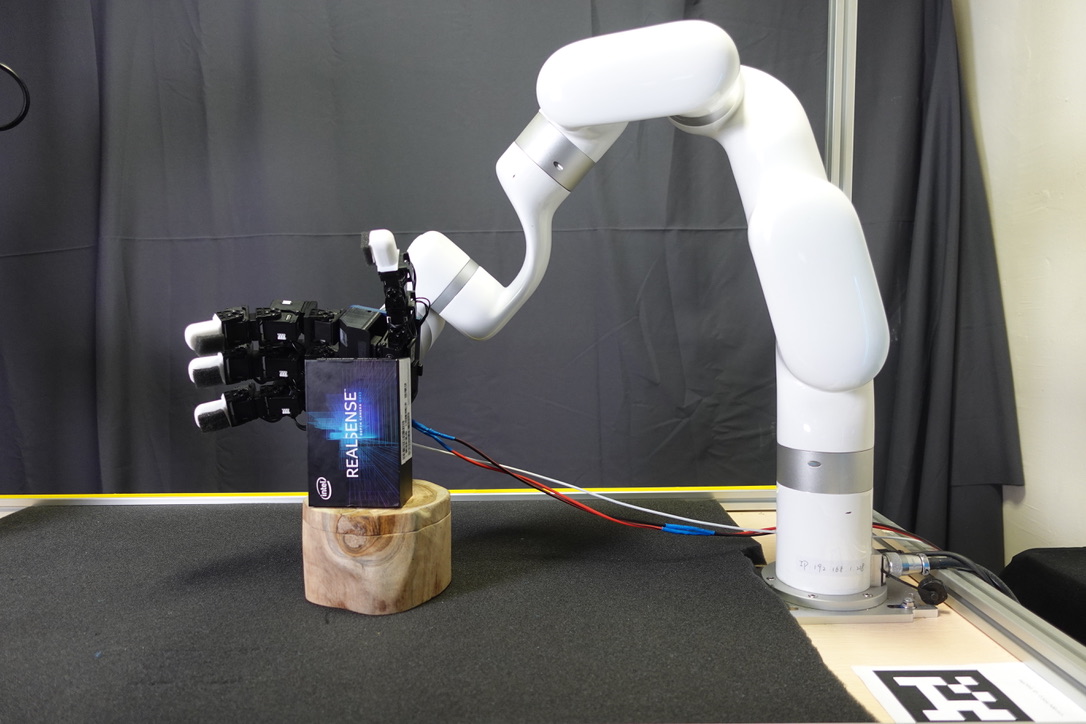}
        \caption{}
    \end{subfigure}
    \caption{Box Rotation Visualization}
    \label{fig:boxrotate}
\end{figure*}

\begin{figure*}[htbp]
    \centering
    \begin{subfigure}[b]{0.18\textwidth}
        \centering
        \includegraphics[width=\textwidth]{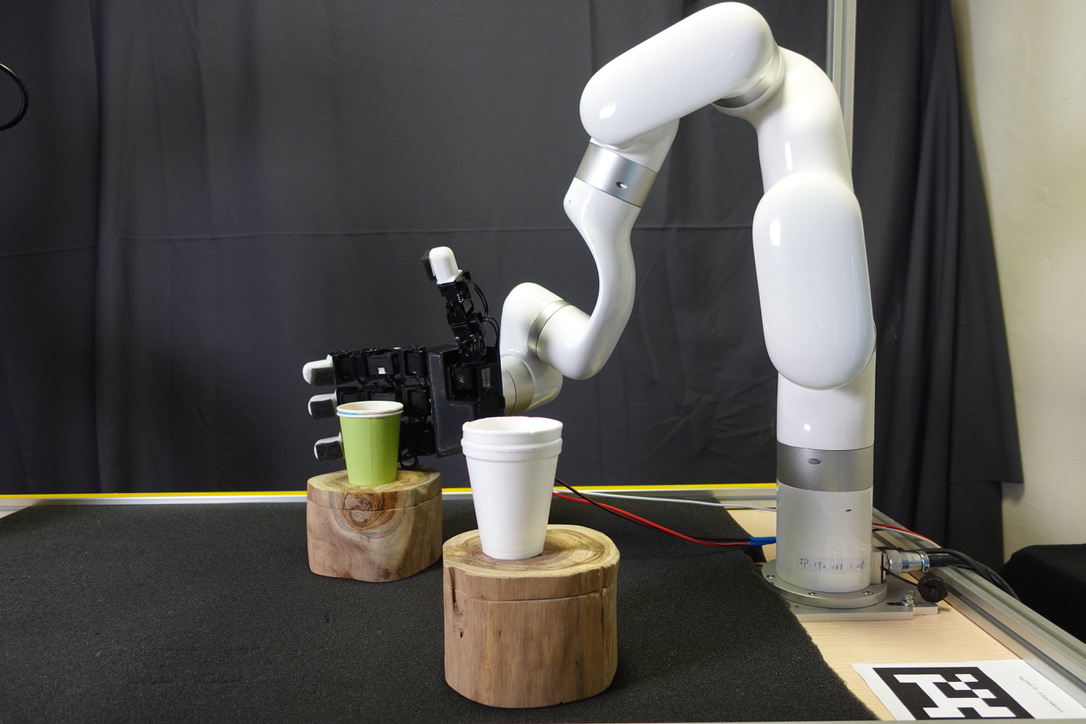}
        \caption{}
    \end{subfigure}
    \begin{subfigure}[b]{0.18\textwidth}
        \centering
        \includegraphics[width=\textwidth]{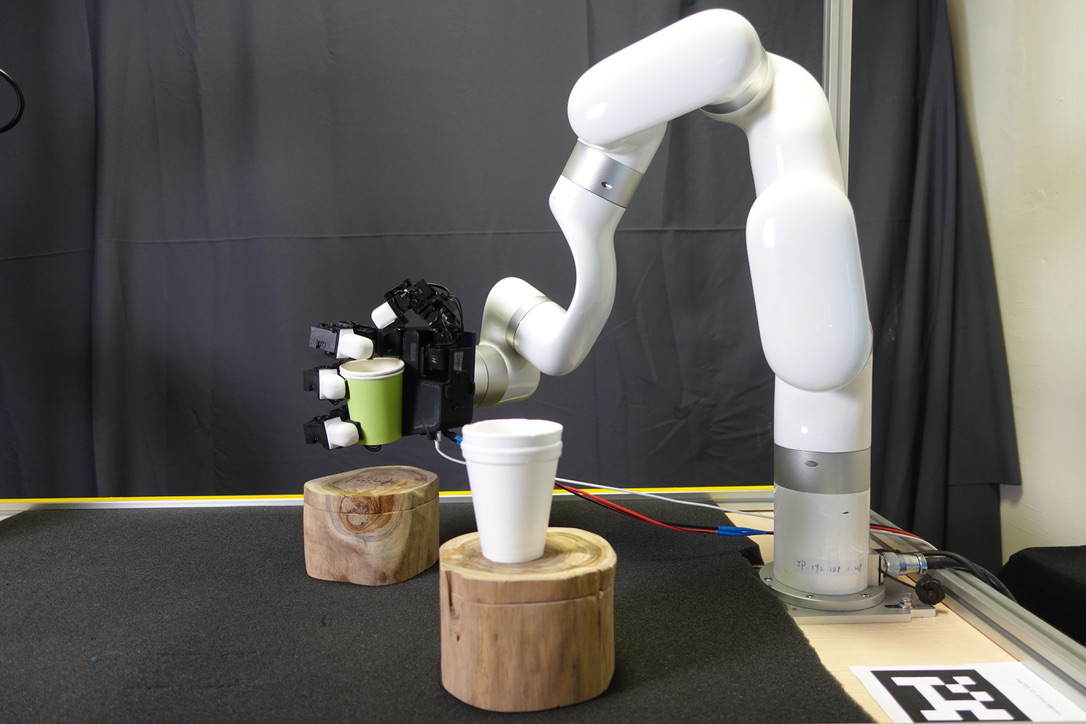}
        \caption{}
    \end{subfigure}
    \begin{subfigure}[b]{0.18\textwidth}
        \centering
        \includegraphics[width=\textwidth]{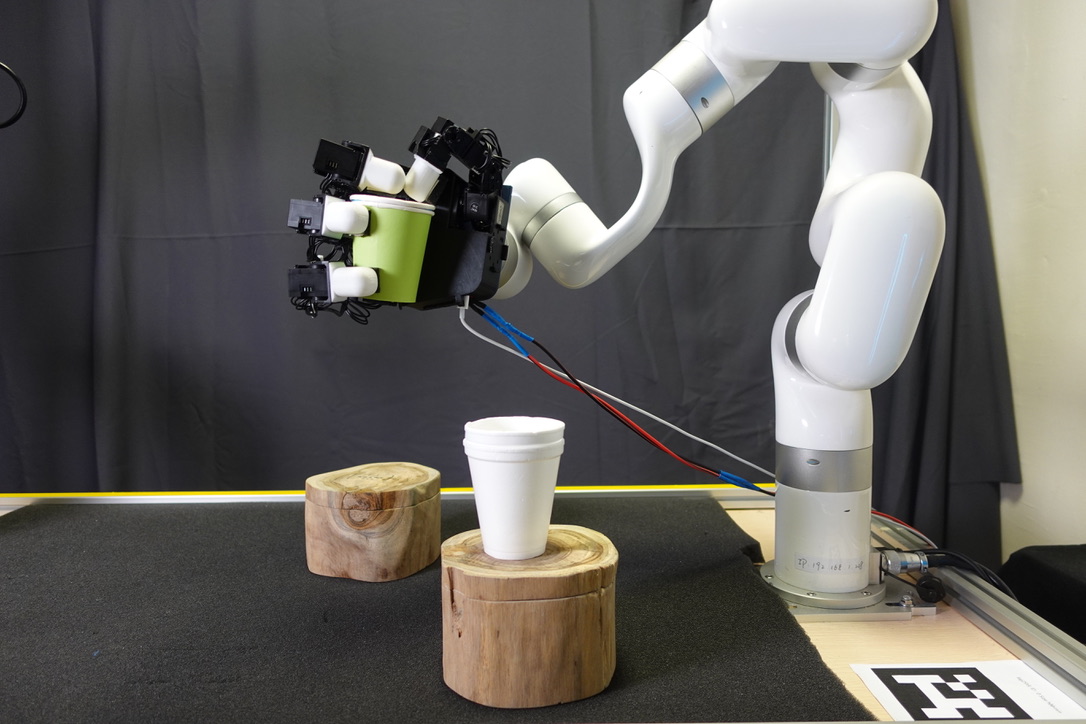}
        \caption{}
    \end{subfigure}
    \begin{subfigure}[b]{0.18\textwidth}
        \centering
        \includegraphics[width=\textwidth]{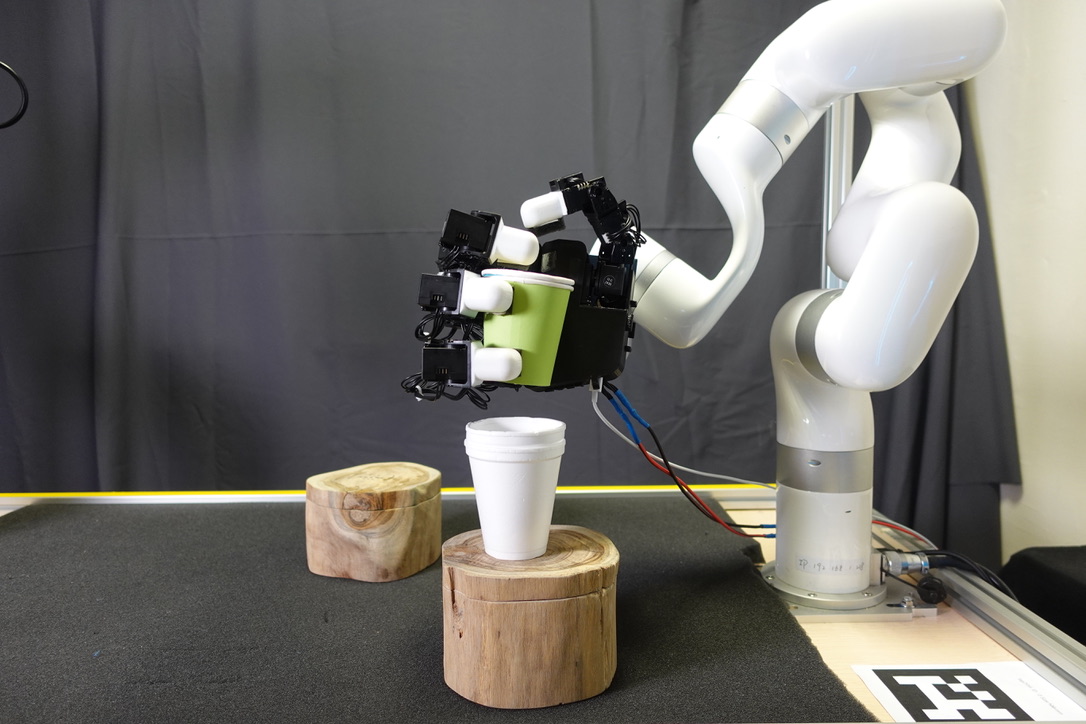}
        \caption{}
    \end{subfigure}
    \begin{subfigure}[b]{0.18\textwidth}
        \centering
        \includegraphics[width=\textwidth]{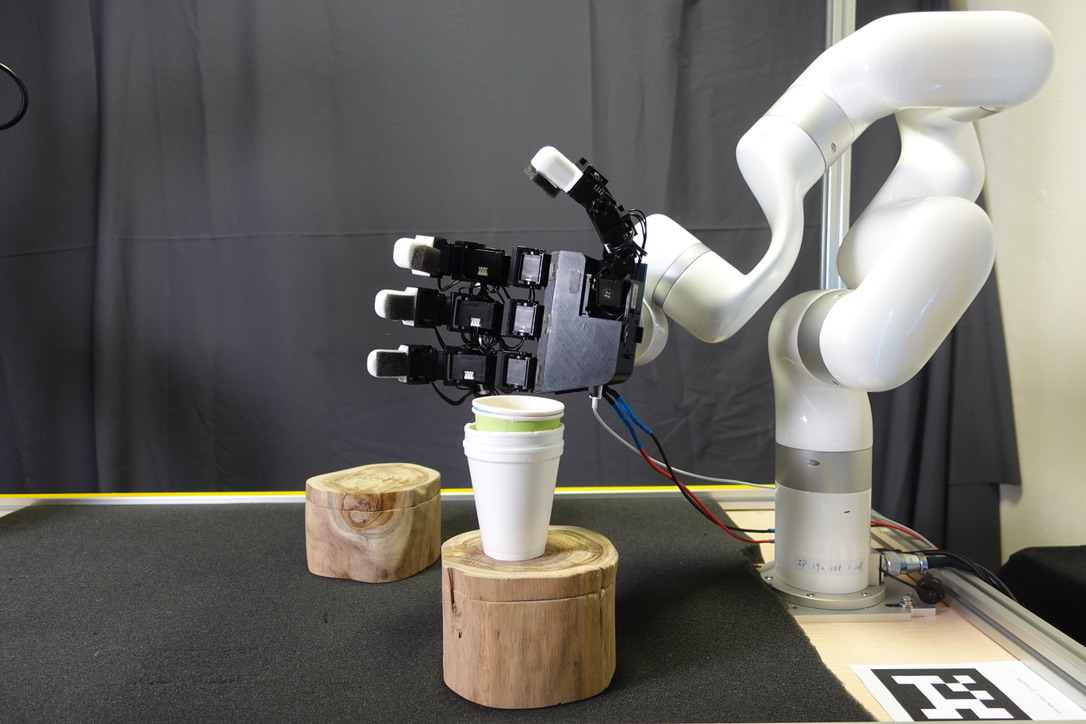}
        \caption{}
    \end{subfigure}
    \caption{Cupstack Visualization}
    \label{fig: cupstack}
\end{figure*}

We visualize the execution process of our five manipulation tasks in Figure \ref{fig: PickPlace}-\ref{fig: cupstack}. They demonstrate the execution sequences of five manipulation tasks: picking and placing a cup, hanging a spoon on a peg, pouring beans between containers, rotating a box, and stacking cups.

\end{document}